\documentclass[preprint,12pt]{elsarticle}

\usepackage{times}
\usepackage{epsfig}
\usepackage{graphicx}
\usepackage{amsmath}
\usepackage{amssymb}
\usepackage{mathtools}
\usepackage{multirow}
\usepackage{comment}
\newcommand\givenbase[1][]{\:#1\lvert\:}
\let\given\givenbase

\usepackage{algorithm}
\usepackage{algorithmic}
\usepackage{subfigure}

\usepackage{booktabs}
\usepackage[table,xcdraw]{xcolor}
\usepackage{adjustbox}
\usepackage{pdflscape}

\usepackage{soul}
\setstcolor{red}

\usepackage{lineno}
\usepackage[colorlinks=true,linkcolor=blue]{hyperref}

\usepackage{rotating}

\journal{Information Fusion}

\begin{document}

\begin{frontmatter}

\title{Unbox the Black-box for the Medical Explainable AI ~\\ via Multi-modal and Multi-centre Data Fusion: ~\\ A Mini-Review, Two Showcases and Beyond}

\author[label1,label2,label3]{Guang Yang\corref{cor1}}
\cortext[cor1]{Corresponding authors.}
\ead{g.yang@imperial.ac.uk}
\author[label4,label5]{Qinghao Ye\corref{cor1}}
\ead{q7ye@ucsd.edu}
\author[label6]{Jun Xia\corref{cor1}}
\ead{xiajun@email.szu.edu.cn}

\address[label1]{National Heart and Lung Institute, Imperial College London, London, UK}
\address[label2]{Royal Brompton Hospital, London, UK}
\address[label3]{Imperial Institute of Advanced Technology, Hangzhou, China}
\address[label4]{Hangzhou Ocean's Smart Boya Co., Ltd}
\address[label5]{University of California, San Diego, La Jolla, California, USA}
\address[label6]{Radiology Department, Shenzhen Second People’s Hospital, Shenzhen, China}





\begin{abstract}
\sloppy
Explainable Artificial Intelligence (XAI) is an emerging research topic of machine learning aimed at \emph{unboxing} how AI systems' \emph{black-box} choices are made. This research field inspects the measures and models involved in decision-making and seeks solutions to explain them explicitly. Many of the machine learning algorithms can not manifest how and why a decision has been cast. This is particularly true of the most popular deep neural network approaches currently in use. Consequently, our confidence in AI systems can be hindered by the lack of explainability in these \emph{black-box} models. The XAI becomes more and more crucial for deep learning powered applications, especially for medical and healthcare studies, although in general these deep neural networks can return an arresting dividend in performance. The insufficient explainability and transparency in most existing AI systems can be one of the major reasons that successful implementation and integration of AI tools into routine clinical practice are uncommon. In this study, we first surveyed the current progress of XAI and in particular its advances in healthcare applications. We then introduced our solutions for XAI leveraging multi-modal and multi-centre data fusion, and subsequently validated in two showcases following real clinical scenarios. Comprehensive quantitative and qualitative analyses can prove the efficacy of our proposed XAI solutions, from which we can envisage successful applications in a broader range of clinical questions.  

\end{abstract}


\end{frontmatter}


\section{Introduction}



Recent years have seen significant advances in the capacity of Artificial Intelligence (AI), which is growing in sophistication, complexity and autonomy. A continuously veritable and explosive growth of data with a rapid iteration of computing hardware advancement provides a \emph{turbo boost} for the development of AI. 

\begin{figure}
\begin{center}
    \includegraphics[width=\linewidth]{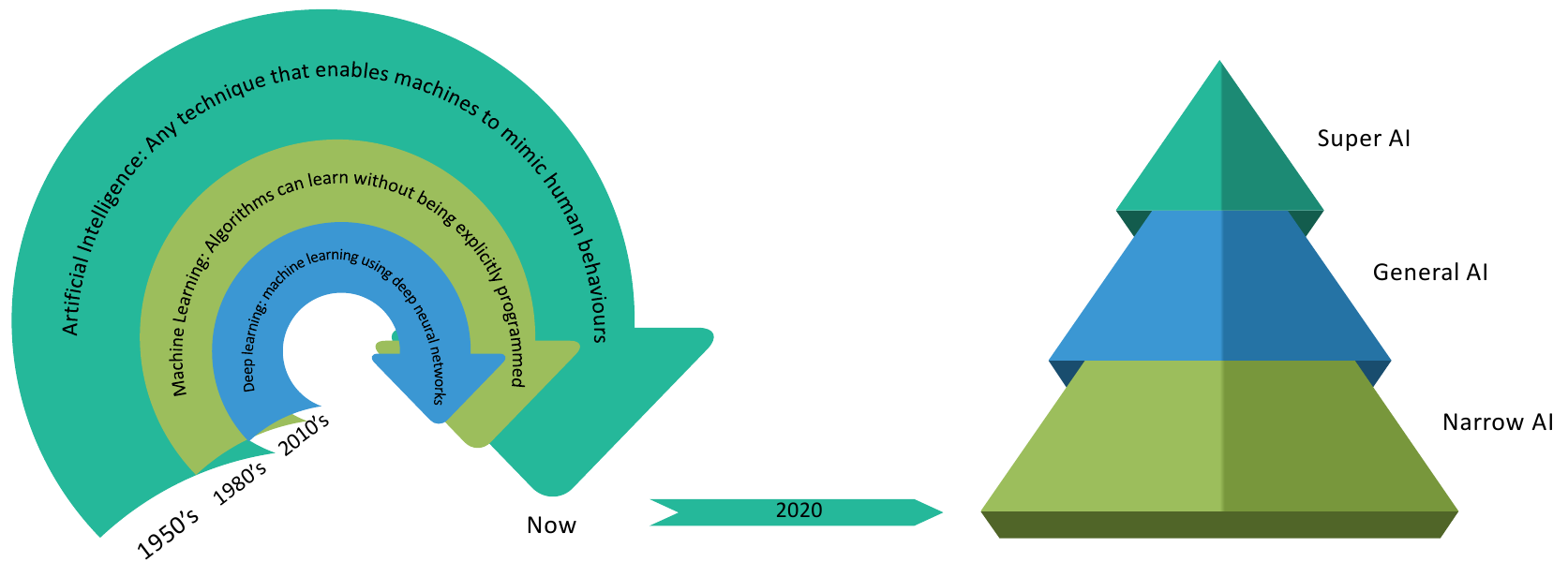} 
    \caption{Left: Terminology and historical timeline of AI, machine learning and deep learning. Right: We are still at the stage of narrow AI, a concept used to describe AI systems that are capable of handling a single or limited task. General AI is the hypothetical wisdom of AI systems capable of comprehending or learning any intelligent activity a human being might perform. Super AI is an AI that exceeds human intelligence and skills.}
    \label{fig:fig1}
\end{center}
\end{figure}

AI is a generic concept and an umbrella term that implies the use of a machine with limited human interference to model intelligent actions. It covers a broad range of research studies from machine intelligence for computer vision, robotics, natural language processing to more theoretical machine learning algorithms design and recently re-branded and thrived \emph{deep learning} development (Figure \ref{fig:fig1}).

\subsection{Born of AI}

AI changes almost every sector globally, e.g., enhancing (digital) healthcare (e.g., making diagnosis more accurate, allowing improved disease prevention), accelerating drug/vaccine development and repurposing, raising agricultural productivity, leading to mitigation and adaptation in climate change, improving the efficiency of manufacturing processes by predictive maintenance, supporting the development of autonomous vehicles and programming more efficient transport networks, and in many other successful applications, which make significant positive socio-economic impact. Besides, AI systems are being deployed in highly-sensitive policy fields, such as facial recognition in the police or recidivism prediction in the criminal justice system, and in areas where diverse social and political forces are presented. Therefore, nowadays, AI systems are incorporated into a wide variety of decision-making processes. As AI systems become integrated into all kinds of decision-making processes, the degree to which people who develop AI, or are subject to an AI-enabled decision, can understand how the resulting decision-making mechanism operates and why a specific decision is reached, has been increasingly debated in science and policy communities.

A collection of innovations, which are typically correlated with human or animal intelligence, is defined as the term “artificial intelligence”. John McCarthy, who coined this term in 1955, described it as “the scientific and technical expertise in the manufacture of intelligent machines", and since then many different definitions have been endowed.

\subsection{Growth of Machine Learning}

Machine learning is a subdivision of AI that helps computer systems to intelligently execute complex tasks. Traditional AI methods, which specify step by step how to address a problem, are normally based on hard-coded rules. Machine learning framework, by contrast, leverages the power of a large amount of data (as examples and not examples) for the identification of characteristics to accomplish a pre-defined task. The framework then learns how the target output will be better obtained. Three primary subdivisions of machine learning algorithms exist:

\begin{itemize}
	\item A machine learning framework, which is trained using labelled data, is generally categorised as supervised machine learning. The labels of the data are grouped into one or more classes at each data point, such as "cats" or "dogs." The supervised machine learning framework exploits the nature from these labelled data (i.e., training data), and forecasts the categories of the new or so called test data.
	
    \item Learning without labels is referred to as unsupervised learning. The aim is to identify the mutual patterns among data points, such as the formation of clusters and allotting data points to these clusters.
    
    \item Reinforcement learning on the other hand is about knowledge learning, i.e., learning from experience. In standard reinforcement learning settings, an agent communicates with its environment, and is given a reward function that it tries to optimise. The purpose of the agent is to understand the effect of its decisions, and discover the best strategies for maximising its rewards during the training and learning procedure.
    
\end{itemize}

It is of note that some hybrid methods, e.g., semi-supervised learning (using partially labelled data) and weakly supervised (using indirect labels), are also under development.

Although not achieving the human-level intelligence often associated with the definition of the general AI, the capacity to learn from knowledge increases the amount and sophistication of tasks that can be tackled by machine learning systems (Figure \ref{fig:fig1}). A wide variety of technologies, many of which people face on a daily basis, are nowadays enabled by rapid developments in machine learning, contributing to current advancements and dispute about the influence of AI in society. Many of the concepts that frame the existing machine learning systems are not new. The mathematical underpinnings of the field date back many decades, and since the 1950s, researchers have developed machine learning algorithms with varying degrees of complexity. In order to forecast results, machine learning requires computers to process a vast volume of data. How systems equipped with machine learning can handle probabilities or uncertainty in decision-making is normally informed by statistical approaches. Statistics, however, often cover areas of research that are not associated with the development of algorithms that can learn to make forecasts or decisions from results. Although several key principles of machine learning are rooted in data science and statistical analysis, some of the complex computational models do not converge with these disciplines naturally. Symbolic approaches, compared to statistical methods, are also used for AI. In order to create interpretations of a problem and to reach a solution, these methods use logic and inference.

\subsection{Boom of Deep Learning}

Deep learning is a relatively recent congregation of approaches that have radically transformed machine learning. Deep learning is not an algorithm per se, but a range of algorithms that implements neural networks with deep layers. These neural networks are so deep that they can only be implemented on computer node clusters --- modern methods of computing --- such as graphics processing units (GPUs), are needed to train them successfully. Deep learning functions very well for vast quantities of data, and it is never too difficult to engineer the functionality even if a problem is complex (for example, due to the unstructured data). When it comes to image detection, natural language processing, and voice recognition, deep learning can always outperform the other types of algorithms. Deep learning assisted disease screening and clinical outcome prediction or automated driving, which were not feasible using previous methods, are well manifested now. Actually, the deeper the neural network with more data loaded for training, the higher accuracy a neural network can produce. The deep learning is very strong, but there are a few disadvantages to it. The reasoning of how deep learning algorithms reach to a certain solution is almost impossible to reveal clearly. Although several tools are now available that can increase insights into the inner workings of the deep learning model, this black-box problem still exists. Deep learning often involves long training cycles, a lot of data and complex hardware specifications, and it is not easy to obtain the specific skills necessary to create a new deep learning approach to tackle a new problem. 

Although acknowledging that AI includes a wide variety of scientific areas, this paper uses the umbrella word 'AI' and much of the recent interest in AI has been motivated by developments in machine learning and deep learning. More importantly, we should realise that there is not one algorithm, though, that will adapt or solve all issues. Success normally depends on the exact problem that needs to be solved and the knowledge available. A hybrid solution is often required to solve the problem, where various algorithms are combined to provide a concrete solution. Each issue involves a detailed analysis into what constitutes the best-fit algorithm. Transparency of the input size, capabilities of the deep neural network and time efficiency should also be taken into consideration, since certain algorithms take a long time to train.

\subsection{Stunt by the Black-box and Promotion of the Explainable AI}

Any of today's deep learning tools are capable of generating extremely reliable outcomes, but they are often highly opaque, if not fully invisible, making it difficult to understand their behaviours. For even skilled experts to completely comprehend these so-called 'black-box' models may be still difficult. As these deep learning tools are applied on a wide scale, researchers and policymakers can challenge whether the precision of a given task outweighs more essential factors in the decision-making procedure.

\begin{figure}
\begin{center}
    \includegraphics[width=\linewidth]{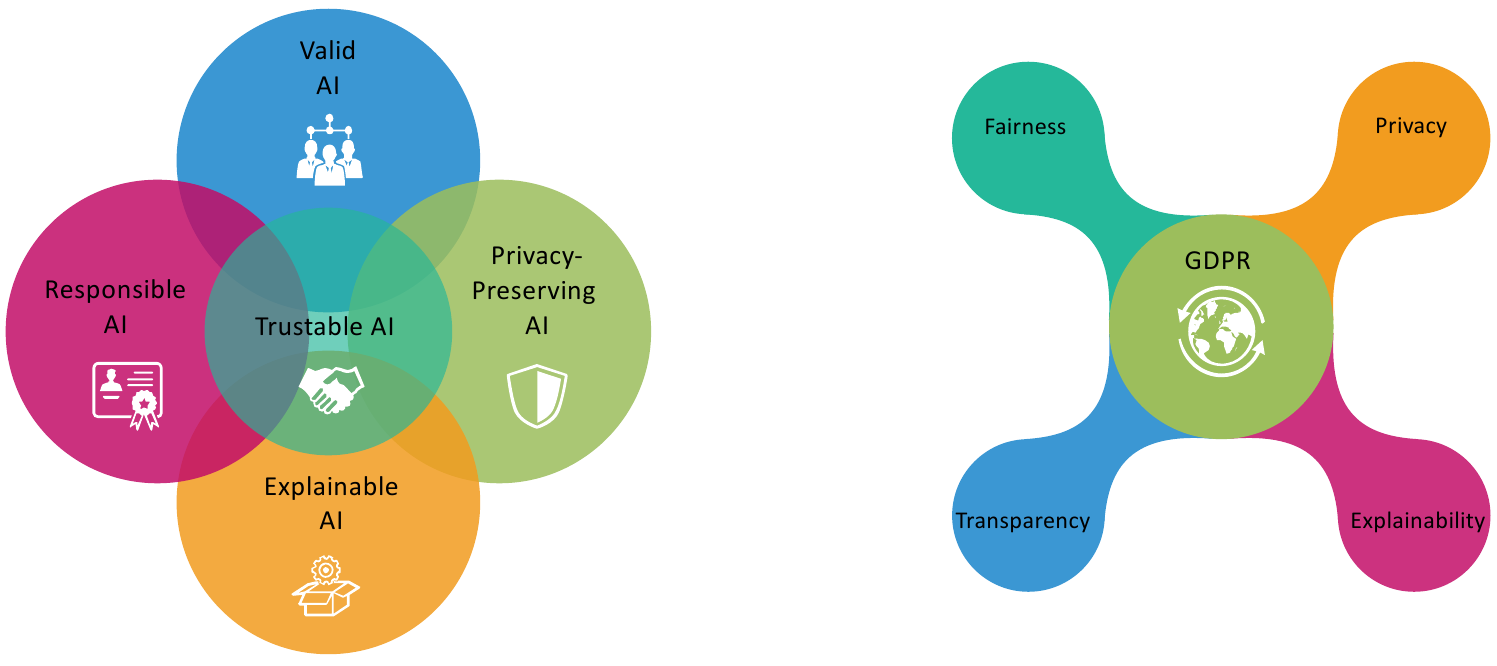} 
    \caption{Left: Trustable AI or Trustworthy AI includes Valid AI, Responsible AI, Privacy-Preserving AI, and Explainable AI (XAI). Right: EU General Data Protection Regulation (GDPR) highlights the Fairness, Privacy, Transparency and Explainability of the AI.}
    \label{fig:fig2}
\end{center}
\end{figure}

As part of attempts to integrate ethical standards into the design and implementation of AI-enabled technologies, policy discussions around the world increasingly involve demands for some form of Trustable AI, which includes Valid AI, Responsible AI, Privacy-Preserving AI, and Explainable AI (XAI), in which the XAI want to address the fundamental question about the rationale of the decision making process including both human level XAI and machine level XAI (Figure \ref{fig:fig2}). For example, in the UK, such calls came from the AI Committee of the House of Lords, which argued that the development of intelligible AI systems is a fundamental requirement if AI will be integrated as a trustworthy tool for our society. In the EU, the High-Level Group on AI has initiated more studies on the pathway towards XAI (Figure \ref{fig:fig2}). Similarly, in the USA, the Defence Advanced Research Projects Agency funds a new research effort aiming at the development of AI with more explainability. These discussions will become more urgent as AI approaches are used to solve problems in a wide variety of complicated policy making areas, as experts increasingly work alongside AI-enabled decision-making tools, for example in clinical studies, and as people more regularly experience AI systems in real life when decisions have a major impact. Meanwhile, research studies in AI continue to progress at a steady pace. XAI is a vigorous area with many on-going studies emerging and several new strategies evolving that make a huge impact on AI development in various ways.

\begin{figure}[!ht]
\begin{center}
    \includegraphics[width=\linewidth]{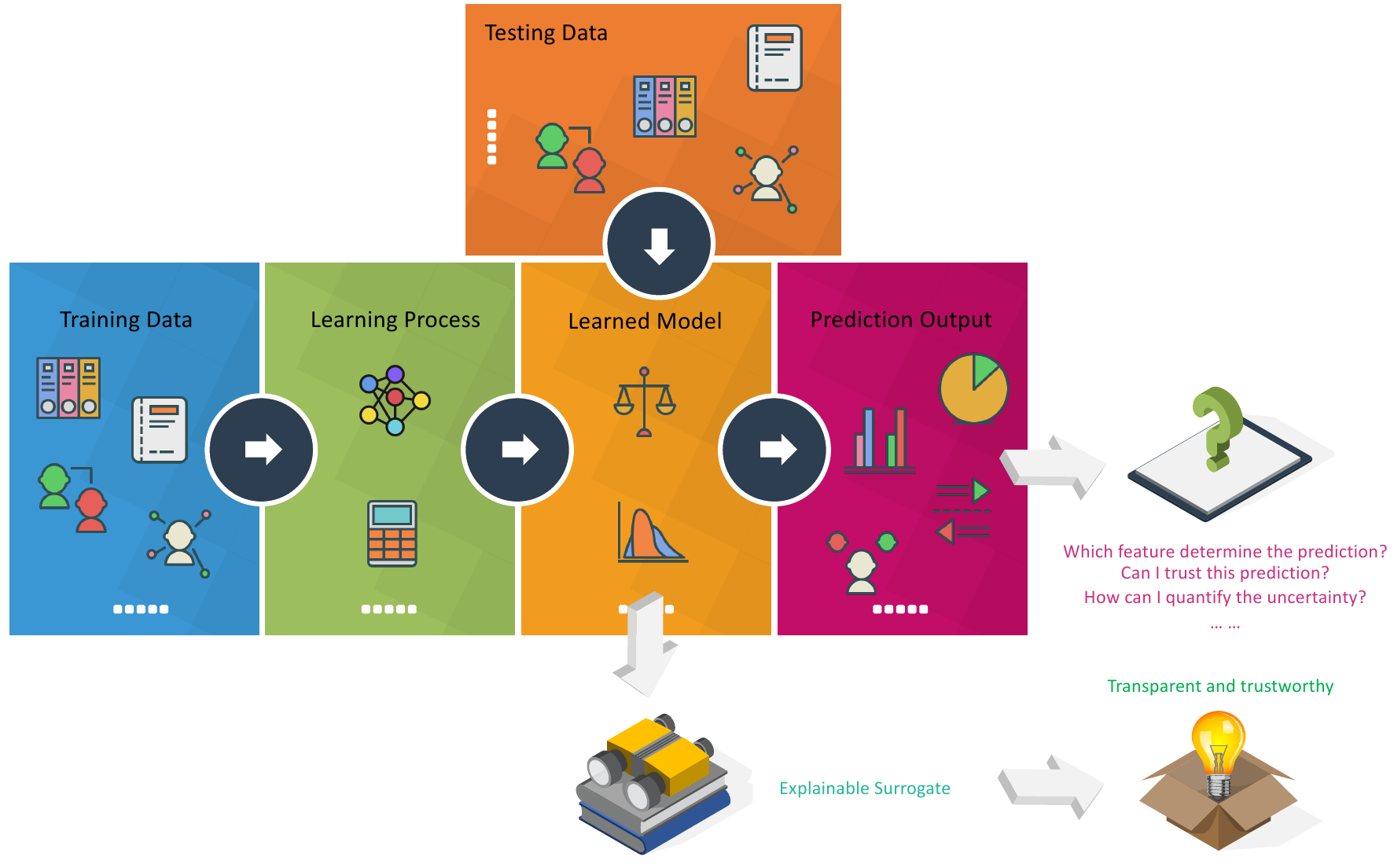} 
    \caption{Schema of the added explainable surrogate module for the normal machine or deep learning procedure that can achieve a more transparent and trustworthy model.}
    \label{fig:fig3}
\end{center}
\end{figure}

\begin{figure}[!htbp]
\begin{center}
    \includegraphics[width=0.8\linewidth]{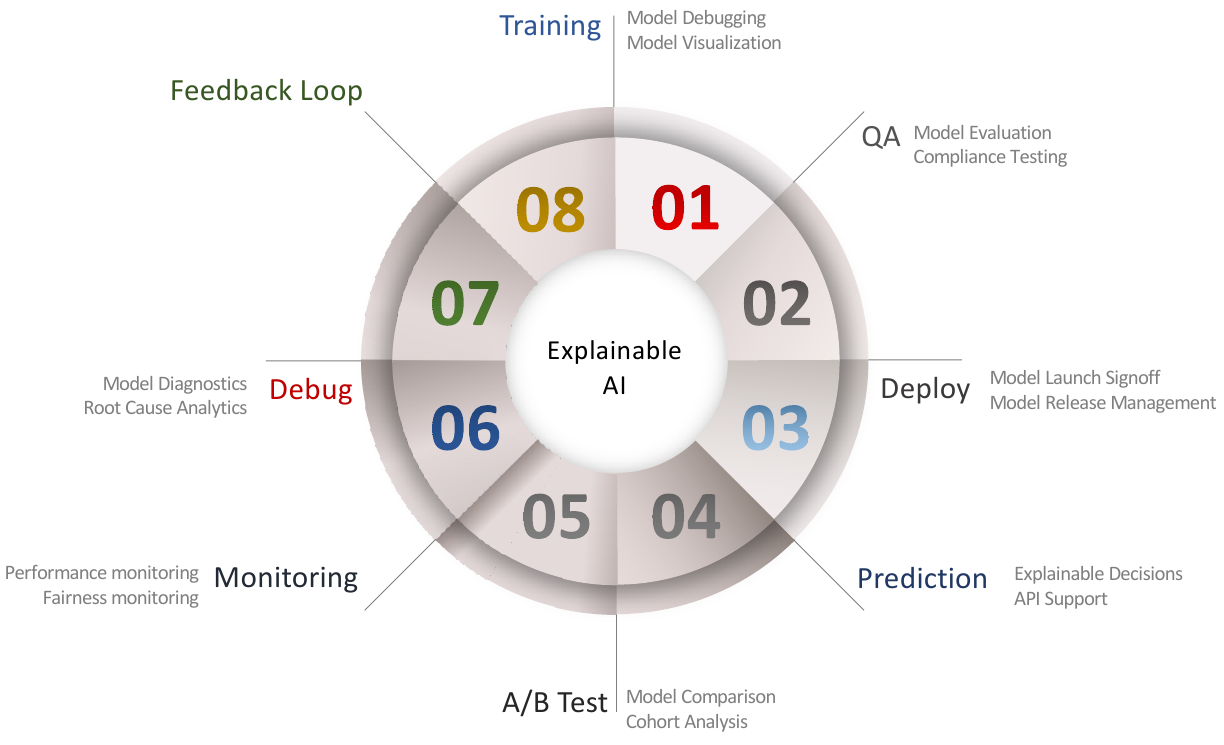} 
    \caption{A typical feedback loop of the XAI development that includes seven steps from training, quality assurance (QA), deployment, prediction, split testing (A/B test), monitoring, and debugging.}
    \label{fig:fig4}
\end{center}
\end{figure}

While the usage of the term is inconsistent, “XAI” refers to a class of systems that have insight into how an AI system makes decisions and predictions. XAI explores the reasoning for the decision-making process, presents the positives and drawbacks of the system, and offers a glimpse of how the system will act in the future. By offering accessible explanations of how AI systems perform their study, XAI can allow researchers to understand the insights that come from research results. For example, in Figure \ref{fig:fig3}, an additional explainable surrogate module can be added to the learnt model to achieve a more transparent and trustworthy model. In other words, for a conventional machine or deep learning model, only generalisation error has been considered while adding an explainable surrogate, both generalisation error and human experience can be considered and a verified prediction can be achieved. In contrast, a learnt black-box model without an explainable surrogate module will cause concerns for the end-users although the performance of the learnt model can be high. Such a black-box model can always cause confusions like ``Why did you do that?", ``Why did you not do that?", ``When do you succeed or fail?", ``How do I correct an error?", and ``Can I trust the prediction?". The XAI powered model, on the other hand, can provide clear and transparent predictions to reassure ``I understand why.", ``I understand why not.", ``I know why you succeed or fail.", ``I know how to correct an error.", and ``I understand, therefore I trust". A typical feedback loop of the XAI development can be found in Figure \ref{fig:fig4}, which includes seven steps from training, quality assurance (QA), deployment, prediction, split testing (A/B test), monitoring, and debugging.   

A variety of terms are used to define certain desired characteristics of an XAI system in research, public, and policy debates, including:

\begin{itemize}
	\item Interpretability: it means a sense of knowing how the AI technology functions.
	
    \item Explainability: it provides an explanation for a wider range of users that how a decision has been drawn.
    
    \item Transparency: it measures the level of accessibility to the data or model.

    \item Justifiability: it indicates an understanding of the case to support a particular outcome.
 
    \item Contestability: it implies how the users can argue against a decision. 
    
\end{itemize}

\begin{figure}[!htb] 
\begin{center}
    \includegraphics[width=0.7\linewidth]{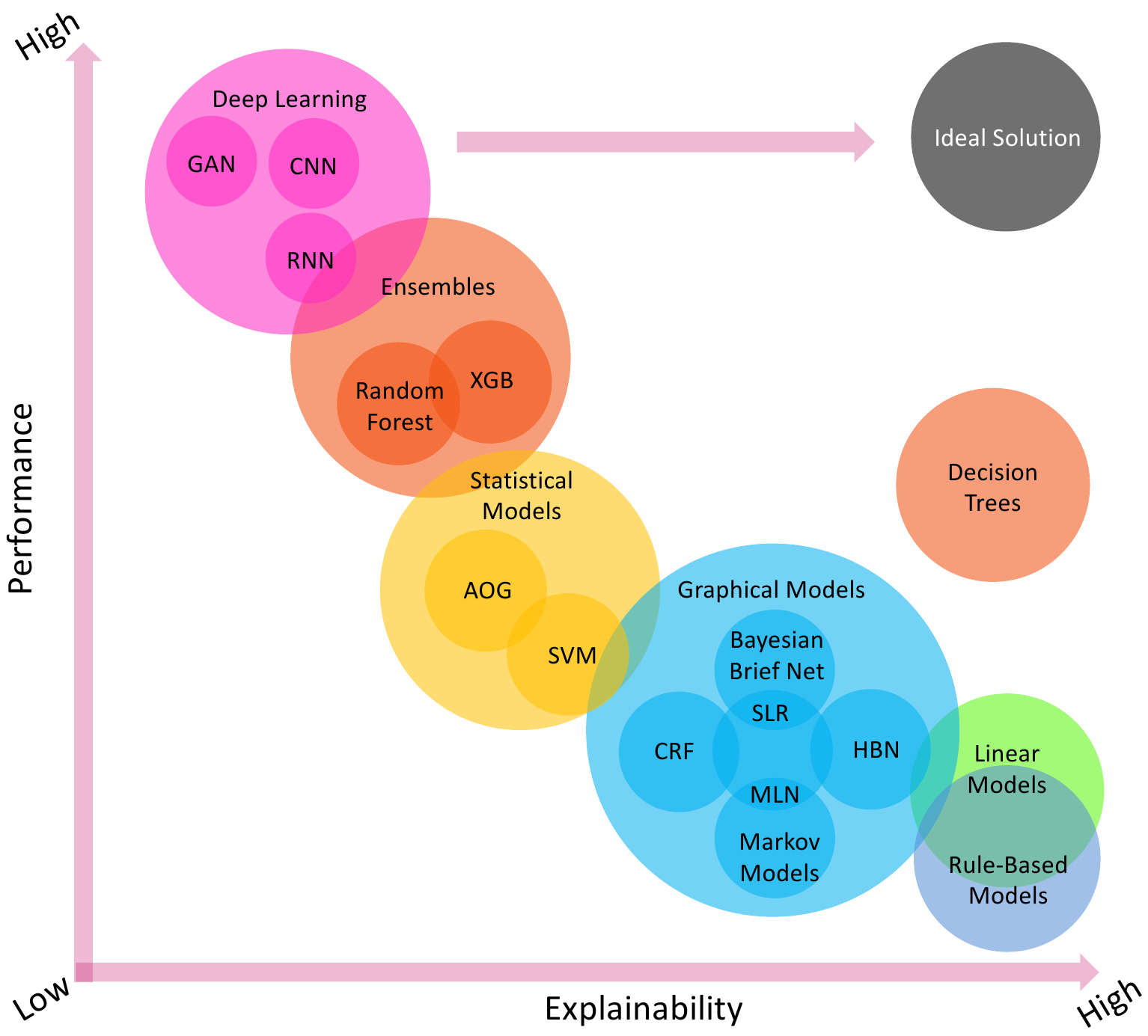} 
    \caption{Model explainability vs. model performance for widely used machine learning and deep learning algorithms. The ideal solution should have both high explainability and high performance. However, existing linear models, rule-based models and decision trees are more transparent, but with lower performance in general. In contrast, complex models, e.g., deep learning and ensembles, manifest higher performance while less explainability can be obtained. HBN: Hierarchical Bayesian Networks; SLR: Simple Linear Regression; CRF: Conditional Random Fields; MLN: Markov Logic Network; SVM: Support Vector Machine; AOG: Stochastic And-Or-Graphs; XGB: XGBoost; CNN: Convolutional Neural Network; RNN: Recurrent Neural Network; and GAN: Generative Adversarial Network.}
    \label{fig:fig5}
\end{center}
\end{figure}

Comprehensive surveys on general XAI can be found elsewhere, e.g., \cite{arrieta2020explainable,adadi2018peeking,samek2019towards,rai2020explainable}; therefore, here we provide an overview of most important concepts of the XAI. Broadly speaking, XAI can be categorised into model-specific or model-agnostic based approaches. Besides, these methods can be classified into local or global methods that can be either intrinsic or post-hoc \cite{arrieta2020explainable}. Essentially, there are many machine learning models that are intrinsically explainable, e.g., linear models, rule-based models and decision trees, which are also known as transparent models or white-box models. However, these relatively simple models may have a relatively lower performance (Figure \ref{fig:fig5}). For more complex models, e.g, support vector machines (SVM), convolutional neural networks (CNN), recurrent neural networks (RNN) and ensemble models, we can design model-specific and post-hoc XAI strategies for each of them. For example, commonly used strategies include explanation by simplification, architecture modification, feature relevance explanation, and visual explanation \cite{arrieta2020explainable}. Clearly these more complex models can achieve better performance while the explainability becomes lower (Figure \ref{fig:fig5}). 

Recently, model-agnostic based approaches attract great attention that rely on a simplified surrogate function to explain the predictions \cite{samek2019towards}. Model-agnostic approaches are not attached to a specific machine learning model. This class of techniques, in other words, distinguishes prediction from explanation. Model-agnostic representations are usually post-hoc that are generally used to explain deep neural networks with interpretable surrogates that can be local or global \cite{adadi2018peeking}. Below is some summary for XAI in more complex deep learning based models.

\subsubsection{Model-Specific Global XAI}

By integrating interpretability constraints into the procedure of deep learning, these model-specific global XAI strategies can improve the understandability of the models. Structural restrictions may include sparsity and monotonicity, where fewer input features are leveraged or the correlation between features and predictions is confined as monotonic). Semantic prior knowledge can also be impelled to restrict the higher-level abstractions derived from the data. For instance, in a CNN based brain tumour detection/classification model using multimodal MRI data fusion, constraints can be imposed by forcing disengaged representations that are recognisable to each MRI modality (e.g., T1, T1 post-contrast and FLAIR), respectively. In doing so, the model can identify crucial information from each MRI modality and distinguish brain tumours and sub-regions into necrotic, more or less infiltrative that can provide vital diagnosis and prognosis information. On the contrary, simple aggregation based information fusion (combining all the multimodal MRI data like a sandwich) would not provide such explainability.

\subsubsection{Model-Specific Local XAI}

In a deep learning model, a model-specific local XAI technique offers an interpretation for a particular instance. Recently, novel attention mechanisms have been proposed to emphasise the importance of different features of the high-dimensional input data to provide an explanation of a representative instance. Consider a deep learning algorithm that encodes an X-ray image into a vector using a CNN and then use an RNN to produce a clinical description for the X-ray image by using the encoded vector. For the RNN, an attention module can be applied to explain to the user what image fragments the model focuses on to produce each substantive term for the clinical description. For example, the attention mechanism will represent the appropriate segments of the image corresponding to the clinical key words derived by the deep learning model when a clinician is baffled to link the clinical key words to the regions of interest in the X-ray image.

\subsubsection{Model-Agnostic Global XAI}

In model-agnostic global XAI, a surrogate representation is developed to approximate an interpretable module for the black-box model. For instance, an interpretable decision tree based model can be used to approximate a more complex deep learning model on how clinical symptoms impact treatment response. A clarification of the relative importance of variables in affecting treatment response to clinical symptoms can be given by the IF-THEN logic of the decision tree. Clinical experts can analyse these variables and are likely to believe the model to the extent that particular symptomatic factors are known to be rational and confounding noises can be accurately removed. Diagnostic methods can also be useful to produce insights into the significance of individual characteristics in the predictions of the model. Partial dependence plots can be leveraged to determine the marginal effects of the chosen characteristics vs. the performance of the forecast, whereas individual conditional expectation can be employed to obtain a granular explanation of how a specific feature affects particular instances and to explore variation in impacts throughout instances. For example, a partial dependency plot can elucidate the role of clinical symptoms in reacting favourably to a particular treatment strategy, as observed by a computer-aided diagnosis system. On the other hand, individual conditional expectation can reveal variability in the treatment response among subgroups of patients.

\subsubsection{Model-Agnostic Local XAI}

For this type of XAI approaches, the aim is to produce model-agnostic explanations for a particular instance or the vicinity of a particular instance. Local Interpretable Model-Agnostic Explanation (LIME) \cite{ribeiro2016lime}, a well-validated tool, can provide an explanation for a complex deep learning model in the neighbourhood of an instance. Consider a deep learning algorithm that classifies a physiological attribute as a high-risk factor for certain diseases or cause of death, for which the clinician requires a post-hoc clarification. The interpretable modules are perturbed to determine how the predictions made by the change of those physiological attributes. For this perturbed dataset, a linear model is learnt with higher weights given to the perturbed instances in the vicinity of the physiological attribute. The most important components of the linear model can indicate the influence of a particular physiological attribute that can suggest a high-risk factor or the contrary can be implied. This can provide comprehensible means for the clinicians to interpret the classifier. 

\section{Related Studies in AI for Healthcare and XAI for Healthcare}

\subsection{AI in Healthcare}

AI attempts to emulate the neural processes of humans, and it introduces a paradigm change to healthcare, driven by growing healthcare data access and rapid development in analytical techniques. We survey briefly the present state of healthcare AI applications and explore their prospects. For a detailed up to date review, the readers can refer to Jiang et al. \cite{jiang2017artificial}, Panch et al. \cite{panch2019inconvenient}, and Yu et al. \cite{yu2018artificial} on general AI techniques for healthcare and Shen et al. \cite{shen2017deep}, Litjens et al. \cite{litjens2017survey} and Ker et al. \cite{ker2017deep} on medical image analysis.

In the medical literature, the effects of AI have been widely debated \cite{dilsizian2014artificial,patel2009coming,jha2016adapting}. Sophisticated algorithms can be developed using AI to 'read' features from a vast amount of healthcare data and then use the knowledge learnt to help clinical practice. To increase its accuracy based on feedback, AI can also be fitted with learning and self-correcting capabilities. By presenting up-to-date medical knowledge from journals, manuals and professional procedures to advise effective patient care, an AI-powered device \cite{strickland2019ibm} will support clinical decision making. Besides, in human clinical practice, an AI system may help to reduce medical and therapeutic mistakes that are unavoidable (i.e., more objective and reproducible) \cite{dilsizian2014artificial,patel2009coming,strickland2019ibm,weingart2000epidemiology, graber2005diagnostic,winters2012diagnostic,lee2013cognitive}. In addition, to help render real-time inferences for health risk warning and health outcome estimation, an AI system can handle valuable knowledge collected from a large patient population \cite{neill2013using}. 

As AI has recently re-emerged into the scientific and public consciousness, AI in healthcare has new breakthroughs and clinical environments are imbued with novel AI-powered technologies at a breakneck pace. Nevertheless, healthcare was described as one of the most exciting application fields for AI. Researchers have suggested and built several systems for clinical decision support since the mid-twentieth century \cite{miller1994medical,musen2014clinical}. Since the 1970s, rule-based methods had many achievements and have been seen to interpret ECGs \cite{kundu2000knowledge}, identify diseases \cite{de1972computer}, choose optimal therapies \cite{shortliffe1975computer}, offer scientific logic explanations \cite{barnett1987dxplain} and assist doctors in developing diagnostic hypotheses and theories in challenging cases of patients \cite{miller1986internist}. Rule-based systems, however, are expensive to develop and can be unstable, since they require clear expressions of decision rules and, like any textbook, require human-authored modifications. Besides, higher-order interactions between various pieces of information written by different specialists are difficult to encode and the efficiency of the structures is constrained by the comprehensiveness of prior medical knowledge \cite{berner1994performance}. To narrow down the appropriate psychological context, prioritise medical theories, and prescribe treatment, it was also difficult to incorporate a method that combines deterministic and probabilistic reasoning procedures \cite{szolovits1978categorical,szolovits1994categorical}.

Recent AI research has leveraged machine learning approaches, which can account for complicated interactions \cite{deo2015machine}, to recognise patterns from the clinical results, in comparison to the first generation of AI programmes, which focused only on the curation of medical information by experts and the formulation of rigorous decision laws. The machine learning algorithm learns to create the correct output for a given input in new instances by evaluating the patterns extracted from all the labelled input-output pairs \cite{yu2016omics}. Supervised machine learning algorithms are programmed to determine the optimal parameters in the models in order to minimise the differences between their training case predictions and the effects observed in these cases, with the hope that the correlations found are generalisable to cases not included in the dataset of training. The model generalisability can be then calculated using the test dataset. For supervised machine learning models, grouping, regression and characterisation of the similarity between instances with similar outcome labels are among the most commonly used tasks. For the unlabelled dataset, unsupervised learning infers the underlying patterns for discovering sub-clusters of the original dataset, for detecting outliers in the data, or for generating low-dimensional data representations. However, it is of note that in a supervised manner, the recognition of low-dimensional representations for labelled dataset may be done more effectively. Machine-learning approaches allow the development of AI applications that promote the exploration of previously unrecognised data patterns without the need to define decision-making rules for each particular task or to account for complicated interactions between input features. Machine learning has therefore been the preferred method for developing AI utilities \cite{deo2015machine,roberts2017biomedical,rogers2020radiomics}.

\begin{figure}[!htb] 
\begin{center}
    \includegraphics[width=0.75\linewidth]{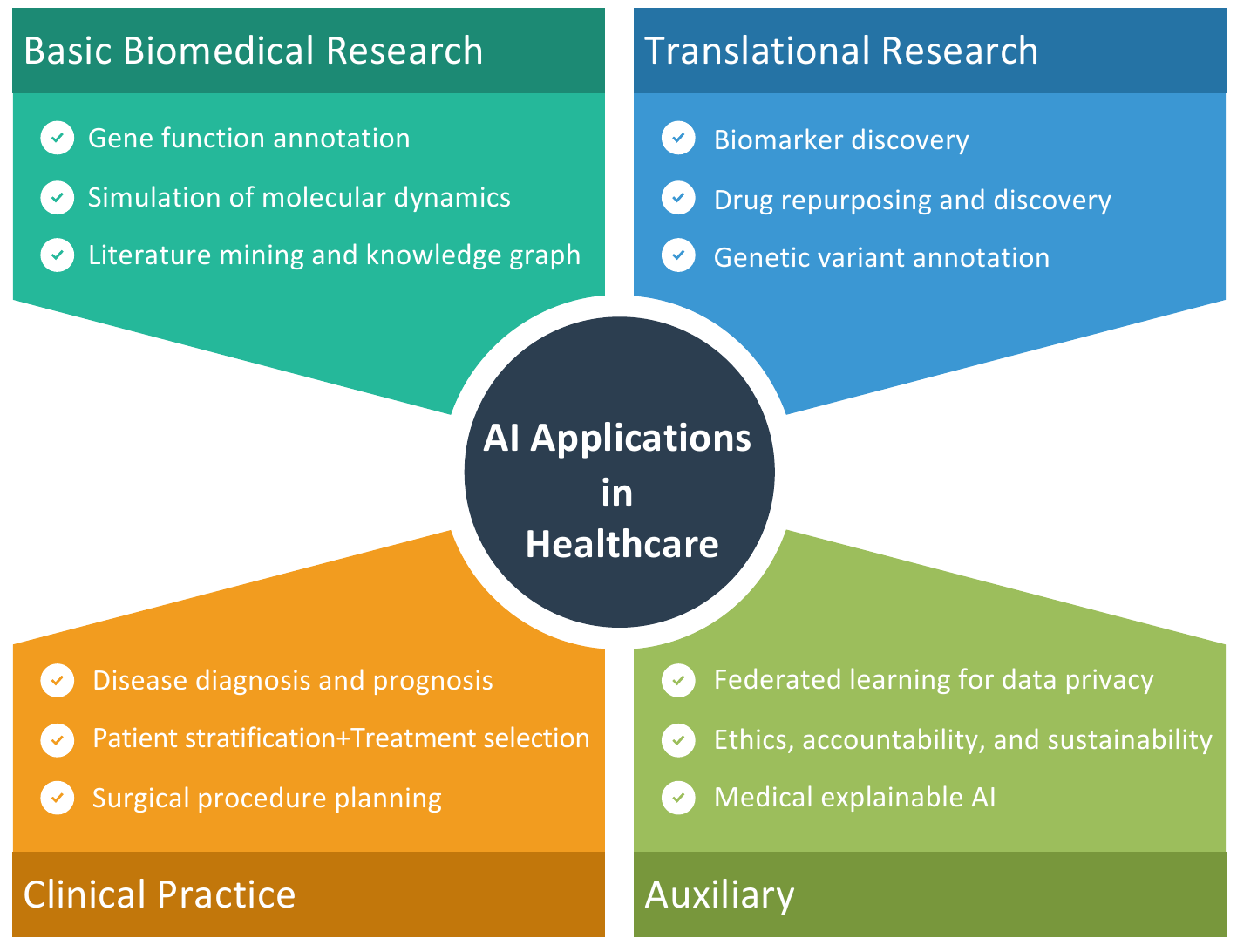} 
    \caption{A non-exhaustive map of the AI in healthcare applications.}
    \label{fig:fig6}
\end{center}
\end{figure}

The recent rebirth of AI has primarily been motivated by the active implementation of deep learning---which includes training a multi-layer artificial neural network (i.e., a deep neural network) on massive datasets---to wide sources of labelled data \cite{goodfellow2016deep}. Existing neural networks are getting deeper and typically have $>$100 layers. Multi-layer neural networks may model complex interactions between input and output, but may also require more data, processing time, or advanced architecture designs to achieve better performance. Modern neural networks commonly have tens of millions to hundreds of millions of parameters and require significant computing resources to perform the model training \cite{yu2018artificial}. Fortunately, recent developments in computer-processor architecture have empowered the computing resources required for deep learning \cite{wang2019benchmarking}. However, in labelled instances, deep-learning algorithms are incredibly 'data hungry.' Huge repositories of medical databases that can be integrated into these algorithms have only recently become readily available, due to the establishment of a range of large-scale research (in particular the Cancer Genome Atlas \cite{tomczak2015cancer} and the UK Biobank \cite{sudlow2015uk}), data collection platforms (e.g., Broad Bioimage Benchmark Collection \cite{ljosa2012annotated} and the Image Data Resources \cite{williams2017image}) and the Health Information Technology for Economic and Clinical Health (HITECH) Act, which has promised to provide financial incentives for the use of electronic health records (EHRs) \cite{desroches2008electronic,hsiao2013office}. In general, deep learning based AI algorithms have been developed for image-based classification \cite{hu2020weakly}, diagnosis \cite{litjens2016deep,zhang2019deep,cao2020multiparameter} and prognosis \cite{cheerla2019deep,roberts2020machine}, genome interpretation \cite{zou2019primer}, biomarker discovery \cite{waldstein2020unbiased,li2020atrial}, monitoring by wearable life-logging devices \cite{lane2015early}, and automated robotic surgery \cite{chen2020deeprobotic} to enhance the digital healthcare \cite{esteva2019guide}. The rapid explosion of AI has given rise to the possibilities of using aggregated health data to generate powerful models that can automate diagnosis and also allow an increasingly precise approach to medicine by tailoring therapies and targeting services with optimal efficacy in a timely and dynamic manner. A non-exhaustive map of possible applications is showing in Figure \ref{fig:fig6}. 

While AI is promising to revolutionise medical practice, several technological obstacles lie ahead. Because deep learning based approaches rely heavily on the availability of vast volumes of high-quality training data, caution must be taken to collect data that is representative of the target patient population. For example, data from various healthcare settings, which include different forms of bias and noise, may cause a model trained in the data of one hospital to fail to generalise to another \cite{obermeyer2016predicting}. Where the diagnostic role has an incomplete inter-expert agreement, it has been shown that consensus diagnostics could greatly boost the efficiency of the training of the deep learning based models \cite{krause2018grader}. In order to manage heterogeneous data, adequate data curation is important. However, achieving a good quality gold standard for identifying the clinical status of the patients requires physicians to review their clinical results independently and maybe repeatedly, which is prohibitively costly at a population scale. A silver standard \cite{rebholz2010calbc} that used natural-language processing methods and diagnostic codes to determine the true status of patients has recently been proposed \cite{kirby2016phekb}. Sophisticated algorithms that can handle the idiosyncrasies and noises of different datasets can improve the efficiency and safety of prediction models in life-and-death decisions.

Most of the recent advancement in neural networks has been limited to well-defined activities that do not require data integration across several modalities. Approaches for the application of deep neural networks to general diagnostics (such as analysis of signs and symptoms, prior medical history, laboratory findings and clinical course) and treatment planning are less simple. While deep learning has been effective in image detection \cite{zhao2019object}, translation \cite{singh2017machine}, speech recognition \cite{amodei2016deep,nassif2019speech}, sound synthesis \cite{purwins2019deep} and even automated neural architecture search \cite{elsken2019neural}, clinical diagnosis and treatment tasks often need more care (e.g., patient interests, beliefs, social support and medical history) than the limited tasks that deep learning can be normally adept. Moreover, it is unknown if transfer learning approaches will be able to translate models learnt from broad non-medical datasets into algorithms for the study of multi-modality clinical datasets. This suggests that more comprehensive data-collection and data-annotation activities are needed to build end-to-end clinical AI programmes.

The design of a computing system for the processing, storage and exchange of EHRs and other critical health data remains a problem \cite{lee2009ethical}. Privacy-preserving approaches, e.g., via federated learning, can allow safe sharing of data or models across cloud providers \cite{narayan2010privacy}. However, the creation of interoperable systems that follow the requirement for the representation of clinical knowledge is important for the broad adoption of such technology \cite{dolin2006hl7}. Deep and seamless incorporation of data across healthcare applications and locations remains questionable and can be inefficient. However, new software interfaces for clinical data are starting to show substantial adoption through several EHR providers, such as Substitutable Medical Applications and Reusable Technologies on the Fast Health Interoperability Resources platform \cite{mandl2012escaping,mandel2016smart}. Most of the previously developed AI in healthcare applications were conducted on retrospective data for the proof of concept \cite{topol2019high}. Prospective research and clinical trials to assess the efficiency of the developed AI systems in clinical environments are necessary to verify the real-world usefulness of these medical AI systems \cite{yu2019framing}. Prospective studies will help recognise the fragility of the AI models in real-world heterogeneous and noisy clinical settings and identify approaches to incorporate medical AI for existing clinical workflows.

AI in medicine would eventually result in safety, legal and ethical challenges \cite{miller2019medical} with respect to medical negligence attributed to complicated decision-making support structures, and have to face the regulation hurdles \cite{challen2019artificial}. If malpractice lawsuits involving medical AI applications occur, the judicial system will continue to provide specific instructions as to which agency is responsible. Health providers with malpractice insurance have to be clear on coverage as health care decisions are taken in part by the AI scheme \cite{yu2018artificial}. With the deployment of automatic AI for particular clinical activities, the criteria for diagnostic, surgical, supporting and paramedical tasks will need to be revised and the functions of healthcare practitioners will begin to change as different AI modules are implemented into the quality of treatment, and the bias needs to be minimised while the patient satisfaction must be maximised \cite{decamp2020latent,esmaeilzadeh2020use}.

\subsection{XAI in Healthcare}

Despite deep learning based AI technologies will usher in a new era of digital healthcare, challenges exist. XAI can play a crucial role, as an auxiliary development (Figure \ref{fig:fig6}), for potentially solving the small sample learning by filter out clinically meaningless features. Moreover, many high-performance deep learning models produce findings that are impossible for unaided humans to understand. While these models can produce better-than-human efficiency, it is not easy to express intuitive interpretations that can justify model findings, define model uncertainties, or derive additional clinical insights from these computational 'black-boxes.' With potentially millions of parameters in the deep learning model, it can be tricky to understand what the model sees in the clinical data, e.g., radiological images \cite{england2019artificial}. For example, research investigation has explicitly stated that being a black box is a "strong limitation" for AI in dermatology since it is not capable of doing a personalised evaluation by a qualified dermatologist that can be used to clarify clinical facts \cite{gomolin2020artificial}. This black-box design poses an obstacle for the validation of the developed AI algorithms. It is necessary to demonstrate that a high-performance deep learning model actually identifies the appropriate area of the image and does not over-emphasise unimportant findings. Recent approaches have been developed to describe AI models including the visualisation methods. Some widely used levers include occlusion maps \cite{zeiler2014visualizing}, salience maps \cite{Simonyan14a}, class activation maps \cite{selvaraju2017grad}, and attention maps \cite{zhang2017mdnet}. Localisation and segmentation algorithms can be more readily interpreted since the output is an image. Model understanding, however, remains much more difficult for deep neural network models trained on non-imaging data other than images that is a current open question for ongoing research efforts \cite{ribeiro2016lime}.

Deep learning-based AI methods have gained popularity in the medical field, with a wide range of work in automatic triage, diagnosis, prognosis, treatment planning and patient management \cite{jiang2017artificial}. We can find many open questions in the medical field that have galvanised clinical trials leveraging deep learning and AI approaches (e.g., from grand-challenge.org). Nevertheless, in the medical field, the issue of interpretability is far from theoretical development. More precisely, it is noted that interpretabilities in the clinical sectors include considerations not recognised in other areas, including risk and responsibilities \cite{croskerry2017diagnosis,panch2019inconvenient}. Life may be at risk as medical responses are made, and leaving those crucial decisions to AI algorithms that without explainabilities and accountabilities will be irresponsible \cite{quinn2020three}. Apart from legal concerns, this is a serious vulnerability that could become disastrous if used with malicious intent. 

As a result, several recent studies \cite{zhang2017mdnet,pmlr-v106-tonekaboni19a,holzinger2019causability} have been devoted to the exploration of explainability in medical AI. More specifically, specific analyses have been investigated, e.g., chest radiography \cite{kallianos2019far}, emotion analysis in medicine \cite{zucco2018explainable}, COVID-19 detection and classification \cite{hu2020weakly}, and the research encourages understanding of the importance of interpretability in the medical field \cite{langlotz2019roadmap}. Besides, the exposition argues \cite{london2019artificial} that a certain degree of opaqueness is appropriate, that is, it would be more important for us to deliver empirically checked reliable findings than to dwell too hard on how to unravel the black-box. It is advised that readers consider these studies first, at least for an overview of interpretability in medical AI. 

An obvious XAI approach has been taken by many researchers is to provide their predictive models with interpretability. These methods depend primarily on maintaining the interpretability of less complicated AI models while improving their performance by techniques of refinement and optimisation. For example, as Figure \ref{fig:fig5} shows, decision tree based methods are normally interpretable, research studies have been done using automated pruning of decision trees for various classifications of illnesses \cite{stiglic2012comprehensive} and accurate decision trees focused on boosting patient stratification \cite{valdes2016mediboost}. However, such model optimisation is not always straightforward and it is not a trivial task. 

\begin{figure}[!htb] 
\begin{center}
    \includegraphics[width=0.8\linewidth]{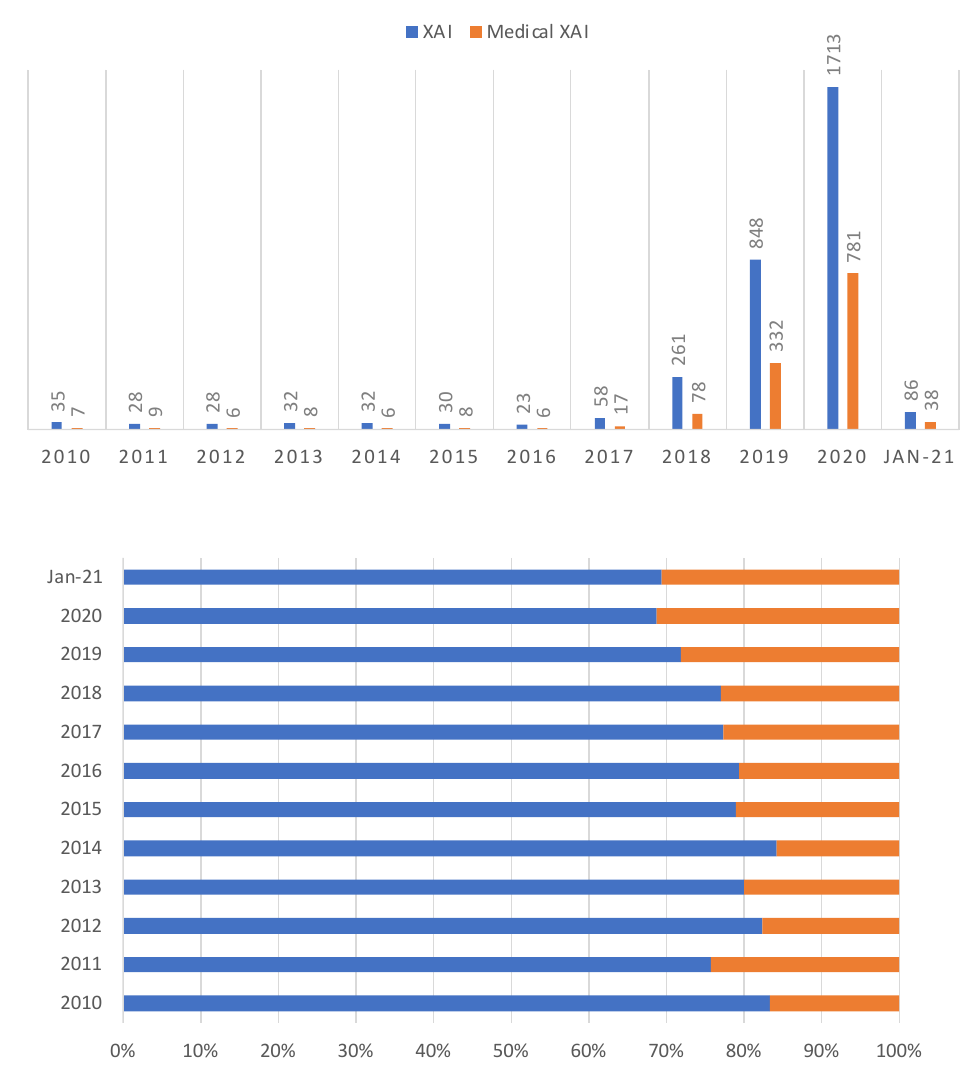} 
    \caption{Publication per year for XAI and medical XAI (top) and percentage for two categories of research (bottom). Data retrieved from Scopus® (Jan 8th, 2021) by using these commands when querying this database---XAI: (ALL(“Explainable AI”) OR ALL(“Interpretable AI”) OR ALL(“Explainable Artificial Intelligence”) OR ALL(“Interpretable Artificial Intelligence”) OR ALL(“XAI”)) AND PUBYEAR = 20XX; Medical XAI: (ALL(“Explainable AI”) OR ALL(“Interpretable AI”) OR ALL(“Explainable Artificial Intelligence”) OR ALL(“Interpretable Artificial Intelligence”) OR ALL(“XAI”)) AND (ALL(“medical”) OR ALL(“medicine”)) AND PUBYEAR = 20XX, in which XX represents the actual year.}
    \label{fig:fig7}
\end{center}
\end{figure}

Previous survey studies on XAI in healthcare can be found elsewhere, e.g., Tjoa and Guan \cite{tjoa2020survey} in medical XAI and Payrovnaziri et al. \cite{payrovnaziri2020explainable} in XAI for EHR. For specific applications, e.g., digital pathology, the readers can refer to Pocevivciute et al. \cite{pocevivciute2020survey} and Tosun et al. \cite{tosun2020histomapr}. The research studies in XAI and medical XAI have been increased exponentially especially after 2018 alongside increasingly development of multimodal clinical information fusion (Figure \ref{fig:fig7}). In this mini-review, we only surveyed the most recent studies that were not covered by previous more comprehensive review studies. In this mini-review, we classified XAI in medicine and healthcare into five categories, which synthesised the approach by Payrovnaziri et al. \cite{payrovnaziri2020explainable}, including (1) XAI via dimension reduction, (2) XAI via feature importance, (3) XAI via attention mechanism, (4) XAI via knowledge distillation, and (5) XAI via surrogate representations (Table \ref{tab:tab1}).

\subsubsection{XAI via Dimension Reduction}

Dimension reduction methods, e.g., using principal component analysis (PCA) \cite{wold1987principal}, independent component analysis (ICA) \cite{comon1994independent}, and Laplacian Eigenmaps \cite{belkin2001laplacian} and other more advanced techniques, are commonly and conventionally used approaches to decipher AI models by representing the most important features. For example, by integrating multi-label k-nearest neighbour and genetic algorithm techniques, Zhang et al. \cite{zhang2015predicting} developed a model for drug side effect estimation based on the optimal dimensions of the input features. Yang et al. \cite{yang2015manifold} proposed a nonlinear dimension reduction method to improve unsupervised classification of the $^1$H MRS brain tumour data and extract the most prominent features using Laplacian Eigenmaps. Zhao and Bolouri \cite{zhao2016object} stratified stage-one lung cancer patients by defining the most insightful examples via a supervised learning scheme. In order to recognise a group of "exemplars" to construct a "dense data matrix," they introduced a hybrid method for dimension reduction by combining pattern recognition with regression analytics. Then they used examples in the final model that are the most predictive for the outcome. Based on domain knowledge, Kim et al. \cite{kim2016opening} developed a deep learning method to extract and rank the most important features based on their weights in the model, and visualised the outcome for predicting cell-type-specific enhancers. To explore the gene pathways and their associations in patients with the brain tumour, Hao et al. \cite{hao2018pasnet} proposed a pathway-associated sparse deep learning method. Bernardini et al. \cite{bernardini2019discovering} used the least absolute shrinkage and selection operator (LASSO) to prompt sparsity for SVMs for the early diagnosis of type 2 diabetes.

Simplifying the information down to a small subset using dimension reduction methods can make the underlying behaviour of the model understandable. Besides, with potentially more stable regularised models, they are less prone to overfitting, which may also be beneficial in general. Nevertheless, the possibility of losing crucial features, which may still be relevant for clinical predictions on a case-by-case basis, can be common and these important features may be neglected unintentionally by the dimensional reduced models.

\subsubsection{XAI via Feature Importance}

Researchers have leveraged the feature importance to explain the characteristics and significance of the extracted features and the correlations among features and between features and the outcomes for providing interpretability for AI models \cite{carvalho2019machine,adadi2018peeking,linardatos2021explainable}. Ge et al. \cite{ge2018interpretable} used feature weights to rank the top ten extracted features to predict mortality of the intensive care unit. Suh et al. \cite{suh2020development} developed a risk calculator model for prostate cancer (PCa) and clinically significant PCa with XAI modules that used Shapley value to determine the feature importance \cite{lundberg2020local}. Sensitivity analysis of the extracted features can represent the feature importance, and essentially the more important features are those for which the output is more sensitive \cite{montavon2018methods}. Eck et al. \cite{eck2017interpretation} defined the most significant features of a microbiota-based diagnosis task by roughly marginalising the features and testing the effect on the model performance.

Shrikumar et al. \cite{shrikumar17a} implemented the Deep Learning Important FeaTures (DeepLIFT)---a backpropagation based approach to realise interpretability. Backpropagation approaches measure the output gradient for input through the backpropagation algorithm to report the significance of the feature. Zuallaert et al. \cite{zuallaert2018splicerover} developed the DeepLIFT based method to create interpretable deep models for splice site prediction by measuring the contribution score for each nucleotide. A recent comparative study of different models of XAI, including DeepLIFT \cite{shrikumar17a}, Guided backpropagation (GBP) \cite{DB15a}, Layer wise relevance propagation (LRP) \cite{bach2015pixel}, SHapley Additive exPlanations (SHAP) \cite{Chen2021} and others, was conducted for ophthalmic diagnosis \cite{singh2020interpretation}.

XAI, by the extraction of feature importance, can not only explain essential feature characteristics, but may also reflect their relative importance to clinical interpretation; however, numerical weights are either not easy to understand or maybe misinterpreted.

\subsubsection{XAI via Attention Mechanism}

The core concept behind the attention mechanism \cite{BahdanauCB14} is that the model "pays attention" only to the parts of the input where the most important information is available. It was originally proposed for tackling the relation extraction task in machine translation and other natural language processing problems. Because certain words are more relevant than others in the relation extraction task, the attention mechanism can assess the importance of the words for the purpose of classification, generating a meaning representation vector. There are various types of attention mechanisms, including global attention, which uses all words to build the context, local attention, which depends only on a subset of words, or self-attention, in which several attention mechanisms are implemented simultaneously, attempting to discover every relation between pairs of words \cite{putelli2019applying}. The attention mechanism has also been shown to contribute to the enhancement of interpretability as well as to technical advances in the field of visualisation \cite{mascharka2018transparency}.

Kaji et al. \cite{kaji2019attention} demonstrated particular occasions when the input features have mostly influenced the predictions of clinical events in ICU patients using the attention mechanism. Shickel et al. \cite{shickel2019deepsofa} presented an interpretable acuity score framework using deep learning and attention-based sequential organ failure assessment that can assess the severity of patients during an ICU stay. Hu et al. \cite{hu2019deephint} provided "mechanistic explanations" for the accurate prediction of HIV genome integration sites. Zhang et al. \cite{zhang2018patient2vec} also built a method to learn how to represent EHR data that could document the relationship between clinical outcomes within each patient. Choi et al. \cite{choi2016retain} implemented the Reverse Time Attention Model (RETAIN), which incorporated two sets of attention weights, one for visit level to capture the effect of each visit and the other at the variable-level. RETAIN was a reverse attention mechanism intended to maintain interpretability, to replicate the actions of clinicians, and to integrate sequential knowledge. Kwon et al. \cite{kwon2018retainvis} proposed a visually interpretable cardiac failure and cataract risk prediction model based on RETAIN (RetainVis). The general intention of these research studies is to improve the interpretability of deep learning models by highlighting particular position(s) within a sequence (e.g., time, visits, DNA) in which those input features can affect the prediction outcome.

Class activation mapping (CAM) \cite{zhou2016learning} method and its variations have been investigated for XAI since 2016, and have been subsequently used for digital healthcare, especially the medical image analysis areas. Lee et al. \cite{lee2019explainable} developed an XAI algorithm for the detection of acute intracranial haemorrhage from small datasets that is one of the most famous studies using CAM. Kim et al. \cite{kimartificial2020} summarised AI based breast ultrasonography analysis with CAM based XAI. Zhao et al. \cite{zhao2018respond} reported a Respond-CAM method that offered a heatmap-based saliency on 3D images obtained from cryo-tomography of cellular electrons. The region where macromolecular complexes were present was marked by the high intensity in the heatmap. Izadyyazdanabadi et al. \cite{izadyyazdanabadi2018weakly} developed a multilayer CAM (MLCAM), which was used for brain tumour localization. Coupling with CNN, Couture et al. \cite{couture2018multiple} proposed a multi-instance aggregation approach to classify breast tumour tissue microarray for various clinical tasks, e.g., histologic subtype classification, and the derived super-pixel maps could highlight the area where the tumour cells were and each mark corresponded to a tumour class. Rajpurkar et al. \cite{rajpurkar2020appendixnet} used Grad-CAM for the diagnosis of appendicitis from a small dataset of CT exams using video pretraining. Porumb et al. \cite{porumb2020precision} combined CNN and RNN for electrocardiogram (ECG) analysis and applied Grad-CAM for the identification of the most relevant heartbeat segments for the hypoglycaemia detection. In Hu et al. \cite{hu2020weakly}, a COVID-19 classification system was implemented with multiscale CAM to highlight the infected areas. By the means of visual interpretability, these saliency maps are recommended. The clinician analysts who examine the AI output can realise that the target is correctly identified by the AI model, rather than mistaking the combination of the object with the surrounding as the object itself.

Attention based XAI methods do not advise the clinical end user specifically of the response, but highlight the areas of greater concern to facilitate easier decision-making. Clinical users can, therefore, be more tolerant of imperfect precision. However, it might not be beneficial to actually offer this knowledge to a clinical end user because of the major concerns, including information overload and warning fatigue. It can potentially be much more frustrating to have areas of attention without clarification about what to do with the findings if the end user is unaware of what the rationale of a highlighted segment is, and therefore the end user can be prone to ignore non-highlighted areas that could also be critical.

\subsubsection{XAI via Knowledge Distillation and Rule Extraction}

Knowledge distillation is one form of the model-specific XAI, which is about eliciting knowledge from a complicated model to a simplified model---enables to train a student model, which is usually explainable, with a teacher model, which is hard to interpret. For example, this can be accomplished by model compression \cite{polino2018model} or tree regularisation \cite{wu2018beyond} or through a coupling approach of model compression and dimension reduction \cite{carvalho2019machine}. Research studies have investigated this kind of technique for several years, e.g., Hinton et al. \cite{Hinton44873}, but has recently been uplifted along with the development of AI interpretability \cite{Hinton46495,yang2020auto,li2020survey}. Rule extraction is another widely used XAI method that is closely associated with knowledge distillation and can have a straightforward application for digital healthcare, for example, decision sets or rule sets have been studied for interpretability \cite{lage2019evaluation} and Model Understanding through Subspace Explanations (MUSE) method \cite{lakkaraju2019faithful} has been developed to describe the projections of the global model by considering the various subgroups of instances defined by user interesting characteristics that also produces explanation in the form of decision sets.

Che et al. \cite{che2016interpretable} introduced an interpretable mimic-learning approach, which is a straightforward knowledge-distillation method that uses gradient-boosting trees to learn interpretable structures and make the baseline model understandable. The approach used the information distilled to construct an interpretable prediction model for the outcome of the ICU, e.g., death, ventilator usage, etc. A rule-based framework that could include an explainable statement of death risk estimation due to pneumonia was introduced by Caruana et al. \cite{caruana2015intelligible}. Letham et al. \cite{letham2015interpretable} also proposed an XAI model named Bayesian rule lists, which offered certain stroke prediction claims. Ming et al. \cite{ming2018rulematrix} developed a visualisation approach to derive rules by approximating a complicated model via model induction at different tasks such as diagnosis of breast cancer and the classification of diabetes. Xiao et al. \cite{xiao2018readmission} built a deep learning model to break the dynamic associations between readmission to hospital and possible risk factors for patients by translating EHR incidents into embedded clinical principles to characterise the general situation of the patients. Classification rules were derived as a way of providing clinicians interpretable representations of the predictive models. Davoodi and Moradi \cite{davoodi2018mortality} developed a rule extraction based XAI technique to predict mortality in ICUs and Das et al. \cite{das2019interpretable} used a similar XAI method for the diagnosis of Alzheimer’s disease. In the LSTM-based breast mass classification, Lee et al. \cite{lee2019generation} incorporated the textual reasoning for interpretability. For the characterisation of stroke and risk prediction, Prentzas et al. \cite{prentzas2019integrating} implemented the argumentation theory for their XAI algorithm training process by extracting decision rules. 

XAI approaches, which rely on knowledge distillation and rule extraction, are theoretically more stable models. The summarised representations of complicated clinical data can provide clinical end-users with the interpretable results intuitively. However, if the interpretation of these XAI results could not be intuitively understood by clinical end-users, then the representations are likely to make it much harder for the end-users to comprehend.

\subsubsection{XAI via Surrogate Representation}

An effective application of XAI in the medical field is the recognition of individual health-related factors that lead to disease prediction using the local interpretable model-agnostic explanation (LIME) method \cite{ribeiro2016lime} that offers explanations for any classifier by approximating the reference model with a surrogate interpretable and "locally faithful" representation. LIME disrupts an instance, produces neighbourhood data, and learns linear models in the neighbourhood to produce explanations \cite{liang2021explaining}. 

Pan et al. \cite{pan2019development} used LIME to analyse the contribution of new instances to forecast central precocious puberty in children. Ghafouri-Fard et al. \cite{ghafouri2019application} have applied a similar approach to diagnose autism spectrum disorder. Kovalev et al. \cite{KOVALEV2020106164} proposed a method named SurvLIME to explain AI base survival models. Meldo et al. \cite{meldo2020natural} used a local post-hoc explanation model, i.e., LIME, to select important features from a special feature representation of the segmented lung suspicious objects. Panigutti et al. \cite{Panigutti2020} developed the "Doctor XAI" system that could predict the readmission, diagnosis and medications order for the patient. Similar to LIME, the implemented system trained a local surrogate model to mimic the black-box behaviour with a rule-based explanation, which can then be mined using a multi-label decision tree. Lauritsen et al. \cite{lauritsen2020explainable} tested an XAI method using Layer-wise Relevance Propagation \cite{montavon2019layer} for the prediction of acute critical illness from EHR.

Surrogate representation is a widely used scheme for XAI; however, the white-box approximation must accurately describe the black-box model to gain trustworthy explanation. If the surrogate models are too complicated or too abstract, the  clinician comprehension might be affected.


\begin{sidewaystable}

\resizebox{0.55\textwidth}{!}{\begin{minipage}{\textwidth}

\begin{tabular}{@{}lllllll@{}}
\toprule
XAI Category           & Reference                                                  & Method                           & Intrinsic/Post-hoc & Local/Global  & Model-specific/Model-agnostic & Application                                                     \\ \midrule
Dimension Reduction    & Zhang et al. \cite{zhang2015predicting}                   & Optimal feature selection        & Intrinsic         & Global        & Model-specific                & Drug side effect estimation                                     \\
                       & Yang et al. \cite{yang2015manifold}                       & Laplacian Eigenmaps              & Intrinsic         & Global        & Model-specific                & Brain tumour classification using MRS                           \\
                       & Zhao and Bolouri \cite{zhao2016object}                    & Cluster analysis and LASSO       & Intrinsic         & Global        & Model-agnostic                & Lung cancer patients stratification                             \\
                       & Kim et al. \cite{kim2016opening}                          & Optimal feature selection        & Intrinsic         & Global        & Model-agnostic                & Cell-type specific enhancers prediction                         \\
                       & Hao et al. \cite{hao2018pasnet}                           & Sparse deep learning             & Intrinsic         & Global        & Model-agnostic                & Long-term survival prediction for glioblastoma multiforme       \\
                       & Bernardini et al. \cite{bernardini2019discovering}        & Sparse-balanced SVM              & Intrinsic         & Global        & Model-agnostic                & Early diagnosis of type 2 diabetes                              \\ \midrule
Feature Importance     & Eck et al. \cite{eck2017interpretation}                   & Feature marginalisation          & Post-hoc           & Global, Local & Model-agnostic                & Gut and skin microbiota/inflammatory bowel diseases diagnosis   \\
                       & Ge et al. \cite{ge2018interpretable}                      & Feature weighting                & Post-hoc           & Global        & Model-agnostic                & ICU mortality prediction (all-cause)                            \\
                       & Zuallaert et al. \cite{zuallaert2018splicerover}          & DeepLIFT                         & Post-hoc           & Global        & Model-agnostic                & Splice site detection                                           \\
                       & Suh et al. \cite{suh2020development}                      & Shapley value                    & Post-hoc           & Global, Local & Model-agnostic                & Decision-supporting for prostate cancer                         \\
                       & Singh et al. \cite{singh2020interpretation}               & DeepLIFT and others              & Post-hoc           & Global, Local & Model-agnostic                & Ophthalmic diagnosis                                            \\ \midrule
Attention Mechanism    & Kwon et al. \cite{kwon2018retainvis}                      & Attention                        & Intrinsic         & Global, Local & Model-specific                & Clinical risk prediction (cardiac failure/cataract)             \\
                       & Zhang et al. \cite{zhang2018patient2vec}                  & Attention                        & Intrinsic         & Local         & Model-specific                & EHR based future hospitalisation prediction                     \\
                       & Choi et al. \cite{choi2016retain}                         & Attention                        & Intrinsic         & Local         & Model-specific                & Heart failure prediction                                        \\
                       & Kaji et al. \cite{kaji2019attention}                      & Attention                        & Intrinsic         & Global, Local & Model-specific                & Predictions of clinical events in ICU                           \\
                       & Shickel et al. \cite{shickel2019deepsofa}                 & Attention                        & Intrinsic         & Global, Local & Model-specific                & Sequential organ failure assessment/in-hospital mortality       \\
                       & Hu et al. \cite{hu2019deephint}                           & Attention                        & Intrinsic         & Local         & Model-specific                & Prediction of HIV genome integration site                       \\ \cmidrule{2-7}
                       & Izadyyazdanabadi et al. \cite{izadyyazdanabadi2018weakly} & MLCAM                            & Intrinsic         & Local         & Model-specific                & Brain tumour localisation                                       \\
                       & Zhao et al. \cite{zhao2018respond}                        & Respond-CAM                      & Intrinsic         & Local         & Model-specific                & Macromolecular complexes                                        \\
                       & Couture et al. \cite{couture2018multiple}                 & Super-pixel maps                 & Intrinsic         & Local         & Model-specific                & Histologic tumour subtype classification                        \\
                       & Lee et al. \cite{lee2019explainable}                      & CAM                              & Intrinsic         & Local         & Model-specific                & Acute intracranial haemorrhage detection                        \\
                       & Kim et al. \cite{kimartificial2020}                       & CAM                              & Intrinsic         & Local         & Model-specific                & Breast neoplasm ultrasonography analysis                        \\
                       & Rajpurkar et al. \cite{rajpurkar2020appendixnet}          & Grad-CAM                         & Intrinsic         & Local         & Model-specific                & Diagnosis of appendicitis                                       \\
                       & Porumb et al. \cite{porumb2020precision}                  & Grad-CAM                         & Intrinsic         & Local         & Model-specific                & ECG based hypoglycaemia detection                               \\
                       & Hu et al. \cite{hu2020weakly}                             & Multiscale CAM                   & Intrinsic         & Local         & Model-specific                & COVID-19 classification                                         \\ \midrule
Knowledge Distillation & Caruana et al. \cite{caruana2015intelligible}             & Rule-based system                & Intrinsic         & Global        & Model-specific                & Prediction of pneumonia risk and 30-day readmission forecast    \\
                       & Letham et al. \cite{letham2015interpretable}              & Bayesian rule lists              & Intrinsic         & Global        & Model-specific                & Stroke prediction                                               \\
                       & Che et al. \cite{che2016interpretable}                    & Mimic learning                   & Post-hoc           & Global, Local & Model-specific                & ICU outcome prediction (acute lung injury)                      \\
                       & Ming et al. \cite{ming2018rulematrix}                     & Visualization of rules           & Post-hoc           & Global        & Model-specific                & Clinical diagnosis and classification (breast cancer, diabetes) \\
                       & Xiao et al. \cite{xiao2018readmission}                    & Complex relationships distilling & Post-hoc           & Global        & Model-specific                & Prediction of the heart failure caused hospital readmission     \\
                       & Davoodi and Moradi \cite{davoodi2018mortality}            & Fuzzy rules                      & Intrinsic         & Global        & Model-specific                & In-hospital mortality prediction (all-cause)                    \\
                       & Lee et al. \cite{lee2019generation}                       & Visual/textual justification     & Post-hoc           & Global, Local & Model-specific                & Breast mass classification                                      \\
                       & Prentzas et al. \cite{prentzas2019integrating}            & Decision rules                   & Intrinsic         & Global        & Model-specific                & Stroke Prediction                                               \\ \midrule
Surrogate Models       & Pan et al. \cite{pan2019development}                      & LIME                             & Post-hoc           & Local         & Model-agnostic                & Forecast of central precocious puberty                          \\
                       & Ghafouri-Fard et al. \cite{ghafouri2019application}       & LIME                             & Post-hoc           & Local         & Model-agnostic                & Autism spectrum disorder diagnosis                              \\
                       & Kovalev et al. \cite{KOVALEV2020106164}                   & LIME                             & Post-hoc           & Local         & Model-agnostic                & Survival models construction                                    \\
                       & Meldo et al. \cite{meldo2020natural}                      & LIME                             & Post-hoc           & Local         & Model-agnostic                & Lung lesion segmentation                                        \\
                       & Panigutti et al. \cite{Panigutti2020}                     & LIME like with rule-based XAI    & Post-hoc           & Local         & Model-agnostic                & Prediction of patient readmission, diagnosis and medications    \\
                       & Lauritsen et al. \cite{lauritsen2020explainable}          & Layer-wise relevance propagation & Post-hoc           & Local         & Model-agnostic                & Prediction of acute critical illness from EHR                   \\ \bottomrule
\end{tabular}

\end{minipage}}
\caption{Summary of various XAI methods in digital healthcare and medicine including their category (XAI via dimension reduction, feature importance, attention mechanism, knowledge distillation, and surrogate representations), reference, key idea, type (Intrinsic or Post-hoc, Local or Global, and Model-specific or Model-agnostic) and specific clinical applications.}
    \label{tab:tab1}

\end{sidewaystable}

\section{Proposed Method}

\subsection{Problem Formulation}

In this study, we have demonstrated two typical but important applications of using XAI, which have been developed for classification and segmentation---two mostly widely discussed problems in medical image analysis and AI-powered digital healthcare. Our developed XAI techniques have been manifested using CT images classification for COVID-19 patients and segmentation for hydrocephalus patients using CT and MRI datasets.

\subsection{XAI for Classification}

\begin{figure}
\begin{center}
    \includegraphics[width=\linewidth]{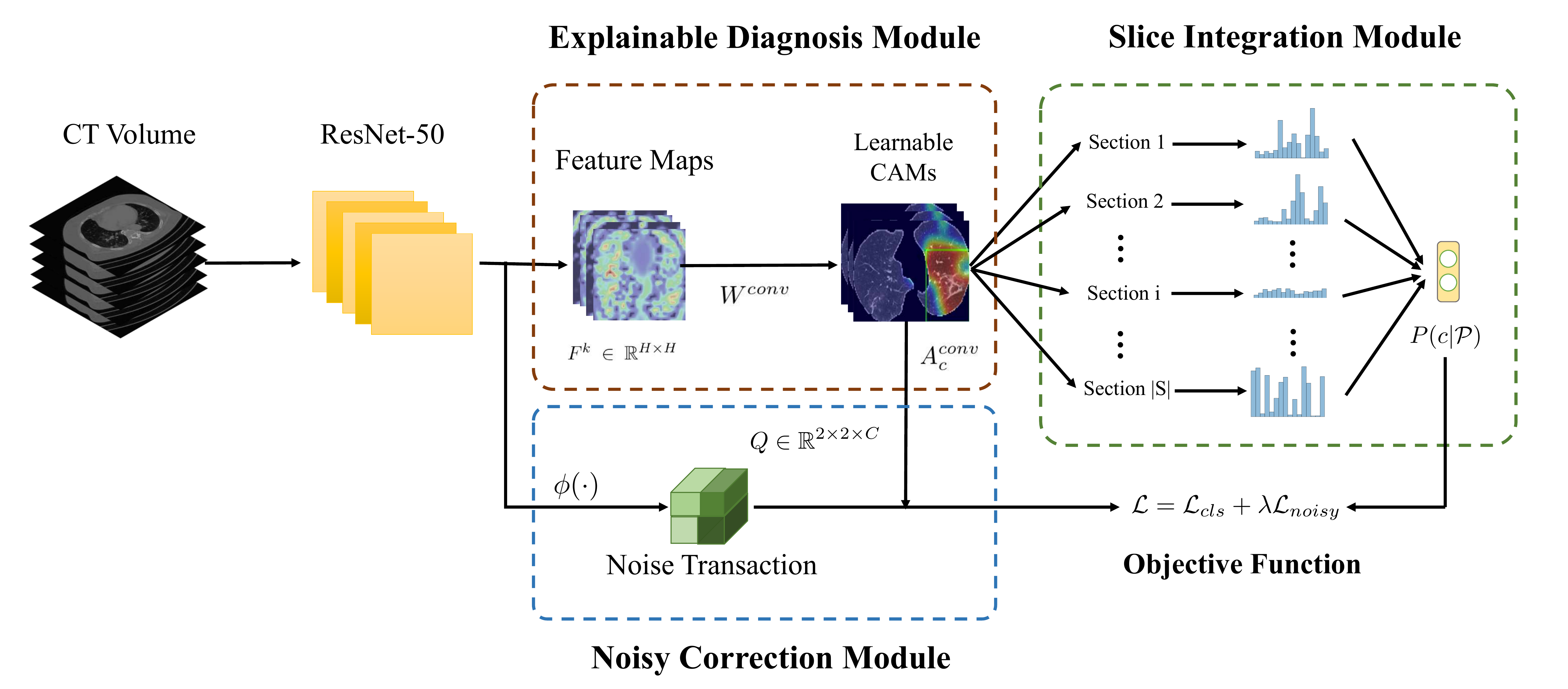} 
    \caption{The overview of our proposed model. $P(c \given S_i)$ denotes the probability of the Section $S_i$, and $P(c \given \mathcal{P})$ represents the probability of the patient who is COVID-19 infected or not. $Q\in \mathbb{R}^{2\times 2 \times C}$ indicates the noise transaction from the probability of the true label $P(y_c \given \mathcal{I})$ to the noise label $P(z_c \given \mathcal{I})$. Besides, $\phi(\cdot)$ is a non-linear feature transformation function, which projects the feature into embedding space.}
    \label{fig:cls_network}
\end{center}
\end{figure}

In this subsection, we provide a practical XAI solution for explainable COVID-19 classification that is capable of alleviating the domain shift problem caused by multicentre data collected for distinguishing COVID-19 patients from other lung diseases using CT images. The main challenge for multicentre data is that hospitals are likely to use different scanning protocols and parameters for CT scanners when collecting data from patients leading to distinct data distribution. Moreover, it can be observed that images obtained from various hospitals are visually different although they are imaging the same organ. If a machine learning model is trained on data from one hospital and tested on the data from another hospital (i.e., another centre), the performance of the model often degrades drastically. Besides, another challenge is that only patient-level annotations are available commonly but image-level labels are not since it would take a large amount of time for radiologists to annotate them \cite{lin2005emergency}. Therefore, we propose a weakly supervised learning based classification model to cope with these two problems. Besides, an explainable diagnosis module in the proposed model can also offer the auxiliary diagnostic information visually for radiologists. The overview of our proposed model is illustrated in Figure \ref{fig:cls_network}.

\subsubsection{Explainable Diagnosis Module (EDM)}

As the predicting process of deep learning models is in a black-box, it is desirable to develop an explainable technique in medical image diagnosis, which provides an explainable auxiliary tool for radiologists. For common practice, CAM can generate the localisation maps for the prediction through the weighted sum of feature maps from the backbone networks such as ResNet \cite{he2016resnet}. Suppose $F^{k} \in \mathbb{R}^{H'\times W'}$ is the $k$-th feature map with the shape of $H' \times W'$, and $W^{fc} \in \mathbb{R}^{K \times C}$, where $K$ is the number of feature maps. Therefore, the class score for class $c$ can be computed as
\begin{align}
    s_c = \sum_{k=1}^K W_{k,c}^{fc} \left( \frac{1}{H'W'} \sum_{i=1}^{H'}\sum_{j=1}^{W'}F_{i,j}^k \right).
\label{eq:cls_cam}
\end{align}
Therefore, the activation map $A_c^{fc}$ for class $c$ can be defined by 
\begin{align}
    (A_c^{fc})_{i,j} = \sum_{k=1}^K W_{k,c}^{fc} F_{i,j}^k.
\end{align}

However, generating CAMs is not an end-to-end process, in which the network should be firstly trained on the dataset and utilises the weights of the last fully connected layer to compute the CAMs, bringing extra computation. To tackle this drawback, in our explainable diagnosis module (EDM), we replace the fully connected layer with a $1\times 1$ convolutional layer of which the weight $W^{conv}$ shares the same mathematical form as $W^{fc}$. So we can reformulate Eq.(\ref{eq:cls_cam}) as
\begin{align}
    s_c =  \frac{1}{H'W'} \sum_{i=1}^{H'}\sum_{j=1}^{W'} \left( \sum_{k=1}^K W_{k,c}^{conv}F_{i,j}^k \right) = \frac{1}{H'W'} \sum_{i=1}^{H'}\sum_{j=1}^{W'} (A_c^{conv})_{i,j},
\end{align}
where $A_c^{conv}$ is the activation map for class $c$ that can be learnt adaptively during the training procedure. The activation map produced by the EDM can not only accurately indicate the importance of the region from CT images and locate the infected parts of the patients, but can also offer the explainable results which are able to account for the prediction.

\subsubsection{Slice Integration Module (SIM)}

Intuitively, each COVID-19 patient case has a different severity. Some patients are severely infected with large lesions, while most of the positive cases can be mild of which only a small portion of the CT volume is infected. Therefore, if we directly apply the patient level annotations as the labels for the image slices, the data would be extremely noisy leading to poor performance as the consequence. To overcome this problem, instead of relying on single images, we propose a slice integration module (SIM) and use the joint distribution of the image slices to model the probability of the patient being infected or not. In our SIM, we assume that the lesions are consecutive and the distribution of the lesion positions is consistent. Therefore, we adopt a section based strategy to handle this problem and ﬁt this into a Multiple Instance Learning (MIL) framework \cite{zhou2004mil}. In the MIL, each sample is regarded as a bag, which is composed of a set of instances. A positive bag contains at least one positive instance, while the negative bag solely consists of negative instances. In our scenario, only patient annotations (bag labels) are provided, and the sections can be regarded as instances in the bags.

Given a patient $\mathcal{P} = [\mathcal{I}_1, \mathcal{I}_2, \cdots, \mathcal{I}_n]$ with $n$ CT slices, we divide them into disjoint sections $\mathcal{P} = \{S_i\}_{i=1}^{|S|}$, where $|S|$ is the total amount of sections for patient $\mathcal{P}$, that is
\begin{align}
    |S| = \max \left(1, \left\lfloor\frac{n}{l_s}\right\rfloor \right).
\end{align}

Here $l_s$ is the section length, which is a designed parameter. Then we integrate the probability of each section as the probability of the patient, that is
\begin{align}
    P(c\givenbase \mathcal{P}) = P(c \givenbase \{S_i\}_{i=1}^{|S|}) = \frac{1}{1+ \prod_{i=1}^{|S|} (\frac{1}{P(c\givenbase S_i)} - 1)}, 
\end{align}
where $P(c\givenbase S_i)$ is the probability of the $i$-th section $S_i$ that belongs to class $c$. By taking the $k$-max probability of the images for each class to compute the section probability, we can mitigate the problem that some slices may contain few infections, which can hinder the prediction for the section. The $k$-max selection method can be formulated as
\begin{align}
    P(c \given S_i) = \sigma\left( \frac{1}{k} \max_{\substack{s^{(j)} \in M}} \sum_{j=1}^k s^{(j)}_c \right), \nonumber \\ s.t. \quad M \subset S_i, |M| = k.
\end{align}
where $s^{(j)}_c$ is the top $j$-th class score of the slice in the $i$-th section for the class $c$, and $\sigma (\cdot)$ represents the Sigmoid function. Then we apply the patient annotations $\textbf{y}$ to compute the classification loss, which can be formulated as
\begin{align}
    \mathcal{L}_{cls} = -\sum_{c = 0}^1 \left[y_c\log P(c \given \mathcal{P}) + (1-y_c)\log (1-P(c \given \mathcal{P})) \right].
\label{eq:loss_cls}
\end{align}

\subsubsection{Noisy Correction Module (NCM)}

In real-world applications, radiologists would only diagnose the disease from one image. Therefore, it is also significant for improving the prediction accuracy on single images. However, the image-level labels are extremely noisy since only patient-level annotations are available. To further alleviate the negative impact of patient-level annotations, we propose a noisy correction module (NCM). Inspired by \cite{bekker2016training}, we model the noise transaction distribution $P(z_c = i \given y_c = j, \mathcal{I})$, which transforms the true posterior distribution $P(y_c \given \mathcal{I})$ to the noisy label distribution $P(z_c \given \mathcal{I})$ by 
\begin{align}
    P(z_c = i \given \mathcal{I}) = \sum_j P(z_c = i \given y_c = j, \mathcal{I}) P(y_c = j \given \mathcal{I}).
\label{eq:noise_prob_simple}
\end{align}

In practice, we estimate the noise transaction distribution $Q^c_{ij} = P(z_c = i \given y_c = j, \mathcal{I})$ for the class $c$ via
\begin{align}
    Q^c_{ij} = P(z_c = i \given y_c = j, \mathcal{I}) = \frac{\exp(w^c_{ij} \phi(\mathcal{I}) + b^c_{ij})}{\sum_i\exp(w^c_{ij} \phi(\mathcal{I}) + b^c_{ij})},
\label{eq:noise_trans}
\end{align}
where $i, j \in \{0,1\}$; $\phi(\cdot)$ is a nonlinear mapping function implemented by convolution layers; $w^c_{ij}$ and $b^c_{ij}$ are the trainable parameters. The noise transaction score $T^c_{ij} = w^c_{ij} \phi(\mathcal{I}) + b^c_{ij}$ represents the confidence score of the transaction from the true label $i$ to the noise label $j$ for the class $c$. Therefore, Eq.(\ref{eq:noise_prob_simple}) can be reformulated as
\begin{align}
    P(z_c = i \given \mathcal{I}) = \sum_j Q^c_{ij} P(y_c = j \given \mathcal{I}).
\label{eq:noise_prob}
\end{align}

By estimating the noisy label distribution $P(z_c \given \mathcal{I})$ for patient $\mathcal{P}$, the noisy classification loss can be computed by
\begin{align}
    \mathcal{L}_{noisy} = -\frac{1}{N} \sum_{n=1}^N\sum_{c = 0}^1 [y_c^n\log P(z_c = 1 \given \mathcal{I}_n) \nonumber \\+ (1-y_c^n)\log P(z_c = 0 \given \mathcal{I}_n)].
\label{eq:loss_noisy}
\end{align}

By combining Eq. (\ref{eq:loss_cls}) and Eq. (\ref{eq:loss_noisy}), we can obtain the total loss function for our XAI solution of an explainable COVID-19 classification, that is
\begin{align}
    \mathcal{L} = \mathcal{L}_{cls} + \lambda \mathcal{L}_{noisy},
\label{eq:loss}
\end{align}
where $\lambda$ is a hyper-parameter to balance the classification loss $\mathcal{L}_{cls}$ and the noisy classification loss $\mathcal{L}_{noisy}$.

\subsection{XAI for Segmentation}

In this subsection, we introduce an XAI model that is applicable for the explainable brain ventricle segmentation using multimodal MRI data acquired from the hydrocephalus patients. Previous methods \cite{qian2017objective, cherukuri2017learning} have conducted experiments using images with a slice thickness of less than 3 mm. This is because the smaller of the image thickness, the more images could be obtained, which helps improve the representation power of the model. However, in a real-world scenario, it is not practical for clinicians to use these models because labelling these image slices is extremely labour-intensive and time-consuming. Therefore, it is more common for the annotations of images with larger slice thicknesses, which are easily available while those images with smaller slice thickness are not. Besides, models trained only on thick-slice images have poor generalisation on thin-slice images. To alleviate these problems, we proposed a thickness agnostic image segmentation model, which can be applicable for both thick-slice and thin-slice images, but only requires the annotations of thick-slice images during the training procedure.

\begin{figure}
\begin{center}
    \includegraphics[width=\linewidth]{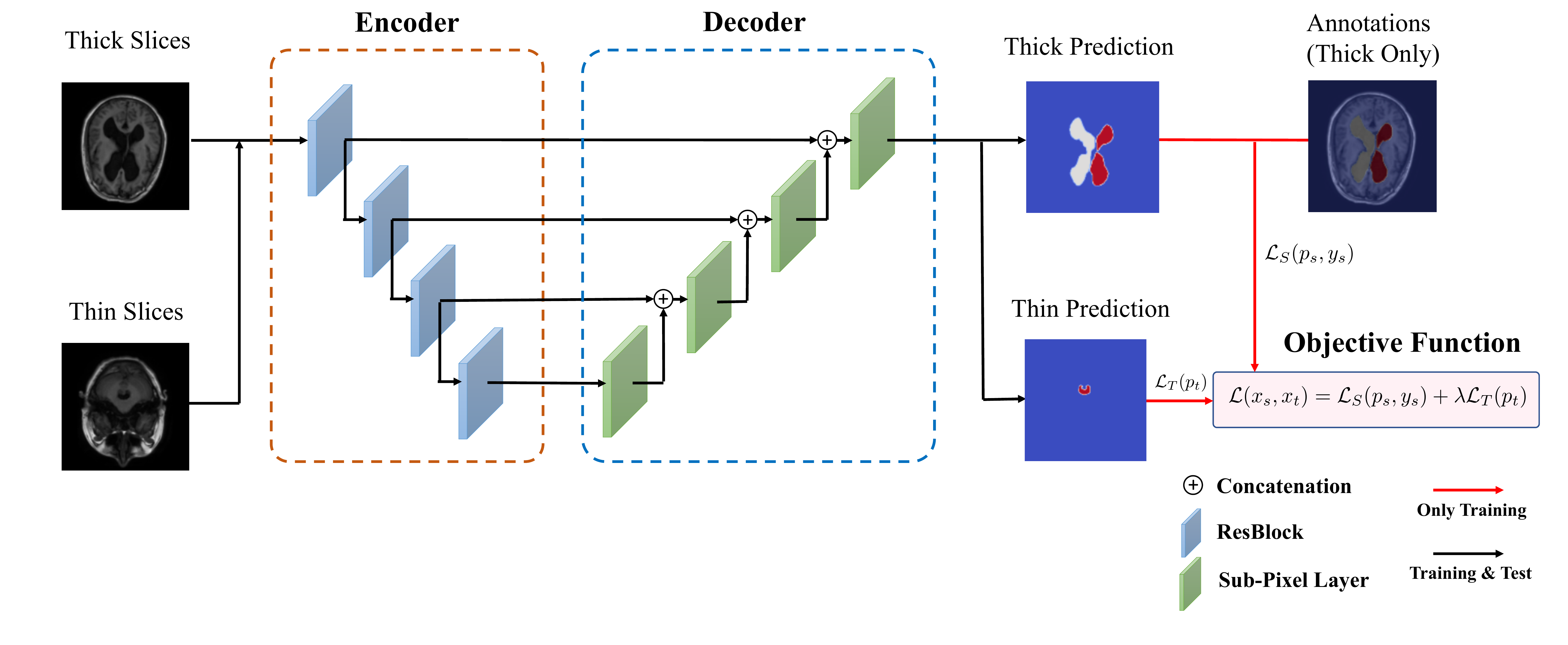} 
    \caption{Overview of our proposed XAI model for explainable segmentation. Here ResBlock represents the residual block proposed in the ResNet \cite{he2016resnet}. }
    \label{fig:seg_network}
\end{center}
\end{figure}

Suppose we have a set of thick-slice images $\mathcal{D}_\mathcal{S} = \{(x_s, y_s) | x_s \in \mathbb{R}^{H\times W \times 3}, y_s \in \mathbb{R}^{H \times W}\}$ and a set of thin-slice images $\mathcal{D}_\mathcal{T} = \{ x_t | x_t \in \mathbb{R}^{H\times W \times 3}\}$. The main idea of our model is to utilise the unlabelled thin-slice images $\mathcal{D}_\mathcal{T}$ to minimise the model performance gap between thick-slice and thin-slice images while a post-hoc XAI can also be developed. 

\subsubsection{Segmentation Network}

With the wide applications of deep learning methods, the encoder-decoder based architectures are usually adopted in automated high accuracy medical image segmentation. The workflow of our proposed segmentation network is illustrated in Figure \ref{fig:seg_network}. Inspired by the U-Net \cite{ronneberger2015unet} model, we replace the original encoder with ResNet-50 \cite{he2016resnet} pre-trained on ImageNet dataset \cite{deng2009imagenet} since it can provide better feature representation for the input images. In addition, the decoder of the U-Net has at least a couple of drawbacks: 1) the increase of low-resolution feature maps can bring a large amount of computational complexity, and 2) interpolation methods \cite{dong2015bilinear} such as bilinear interpolation and bicubic interpolation do not bring extra information to improve the segmentation. Instead, the decoder of our model adopts sub-pixel convolution for constructing segmentation results. The sub-pixel convolution can be represented as 
\begin{align}
    F^{L} = SP(W_L * F^{L-1} + b_L),
\end{align}
where $SP(\cdot)$ operator transforms and arranges a tensor shaped in $H \times W \times C \times r^2$ into a a tensor with the shape of $rH \times rW \times C$, and $r$ is the scaling factor. $F^{L-1}$ and $F^L$ are the input feature maps and output feature maps. $W_L$ and $b_L$ are the parameters of the sub-pixel convolution operators for the layer $L$.

\subsubsection{Multimodal Training}

As aforementioned, the thick-slice images with annotations are available. Therefore, in order to minimise the performance gap between thick-slice images and thin-slice images. We apply a multimodal training procedure to jointly optimise for both types of images. Overall, the objective function of our proposed multimodal training can be computed as
\begin{align}
    \mathcal{L}(x_t, x_s) = \mathcal{L}_\mathcal{S}(p_s, y_s) + \beta \mathcal{L}_\mathcal{T}(p_t),
\end{align}
where $\beta$ is a hyper-parameter for weighting the impact of $\mathcal{L}_\mathcal{S}$ and $\mathcal{L}_\mathcal{T}$. $p_s$ and $p_t$ are the prediction of the segmentation probability maps shaped in $H\times W\times C$ for thick-slice images and thin-slices images, respectively. In particular, $\mathcal{L}_\mathcal{S}$ is the cross-entropy loss defined as follows
\begin{align}
    \mathcal{L}_\mathcal{S} (p_s, y_s) = -\frac{1}{HWC} \sum_{n=1}^{HW} \sum_{c=1}^C y_s^{n,c} \log p_s^{n,c}.
\end{align}

For the unlabelled thin-slice images, we assume that $\mathcal{L}_\mathcal{T}$ can push the features away from the decision boundary of the feature distributions of the thick-slice images, thus achieving distribution alignment. Besides, according to \cite{grandvalet2005semi}, minimising the distance between the prediction distribution $p$ and the uniform distribution $\mathcal{U} = \frac{1}{C}$ can diminish the uncertainty of the prediction. To measure the distance of these two distributions, the objective function $\mathcal{L}_\mathcal{T}$ can be modelled by the $f$-divergence, that is
\begin{align}
    \mathcal{L}_\mathcal{T} (p_t) = -\frac{1}{HWC} \sum_{n=1}^{HW} \sum_{c=1}^C D_f(p_t^{n,c} ||  \mathcal{U}) = -\frac{1}{HWC} \sum_{n=1}^{HW} \sum_{c=1}^Cf(Cp_t^{n,c}).
\end{align}

Most existing methods \cite{grandvalet2005semi, vu2019advent} tend to choose $f(x) = x\log x$, which is alternatively named as KL-divergence. However, one of the main obstacle is that when adopting $f(x) = x\log x$, the gradient of $\mathcal{L}_\mathcal{T}$ would be extremely imbalanced. To be more specific, it can assign a large gradient to the easily classified samples, while assigning a small gradient to hardly classified samples. Therefore, in order to mitigate the unbalancing problem during the optimisation, we incorporate Pearson $\chi^2$-divergence (i.e., $f(x) = x^2-1$) rather than using the KL-divergence for $\mathcal{L}_\mathcal{T}$, that is 
\begin{align}
    \mathcal{L}_\mathcal{T} (p_t) = -\frac{C}{HW} \sum_{n=1}^{HW} \sum_{c=1}^C(p_t^{n,c})^2.
\end{align}

After applying the Pearson $\chi^2$-divergence, the gradient imbalanced issue can be mitigated since the slope of the gradient is constant, which can be verified by taking the second order derivative of $\mathcal{L}_\mathcal{T}$.

During the training procedure, $\mathcal{L}(x_t, x_s)$ is optimised alternatively for both thick-slice and thin-slice images.

\subsubsection{Latent Space Explanation}

Once the model is trained using multimodal datasets, the performance of the network can be quantitatively evaluated by volumetric or regional overlapping metrics, e.g., Dice scores. However, the relation between network performance and input samples remains unclear. In order to provide information about the characteristics of data and their effect on model performance, through which users can set their expectations accordingly, we investigate the feature space and their correlation with the model performance. For feature space visualisation, we extract the outputs of the encoder module of our model, and then decompose them into a two-dimensional space via Principal Component Analysis (PCA). For estimating the whole space, we use a multi-layer perceptron to fit the decomposed samples and their corresponding Dice scores, which can provide an understanding of Dice scores for particular regions of interests in the latent space where there is no data available. Therefore, through analysing the characteristics of the samples in the latent space, we can retrieve the information about the relationships between samples and their prediction power.

\subsection{Implementation Details}
For both our classification and segmentation tasks, we used ResNet-50 \cite{he2016resnet} as the backbone network pre-trained on ImageNet \cite{deng2009imagenet}. For classification, we resized these images into a spatial resolution of $224 \times 224$. During the training procedure, we set $\lambda = 1 \times 10^{-4}$, the dropout rate as $0.7$, and the $L_2$ weight decay coefficient as $1\times 10^{-5}$. Besides, $l_s$ was set to 16, and $k$ was set to 8 for the sake of computing the patient-level probability. For segmentation, we set $\beta = 1 \times 10^{-2}$ for balancing the impact of supervised loss and unsupervised loss. During the training, Adam \cite{kingma2014adam} optimiser was utilised with a learning rate $1 \times 10^{-3}$. The training procedure is terminated after 4,000 iterations with batch size 8. All of the experiments were conducted on a workstation with 4 NVIDIA RTX GPUs using PyTorch framework with version 1.5. 

\section{Experimental Settings and Results}

\subsection{Showcase I: Classification for COVID-19} 

\subsubsection{Datasets}
\begin{figure}[!ht]
\begin{center}
    \subfigure[Patient-level Statistics]{
        \includegraphics[width=0.9\linewidth]{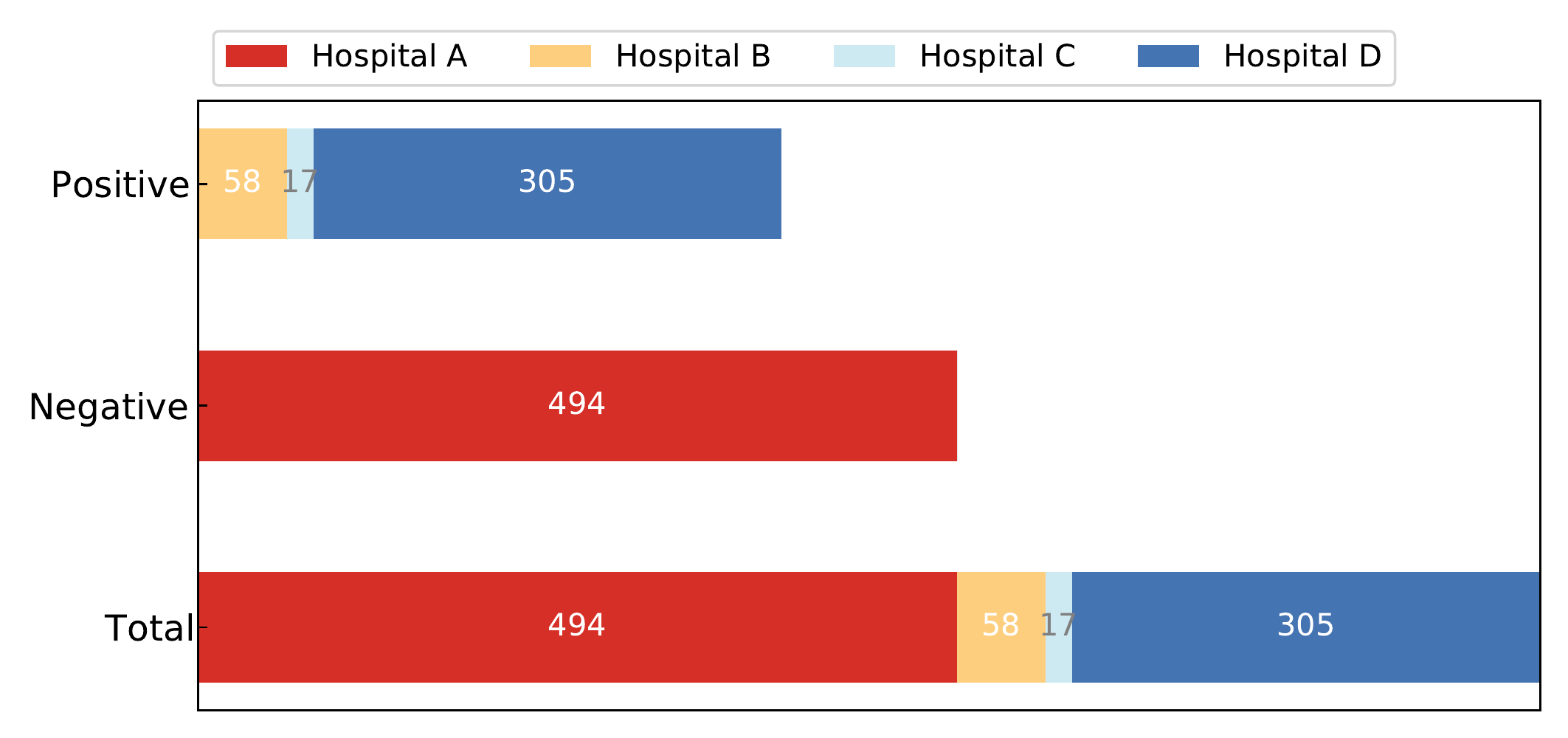} 
    }
    \subfigure[Image-level Statistics]{
        \includegraphics[width=0.9\linewidth]{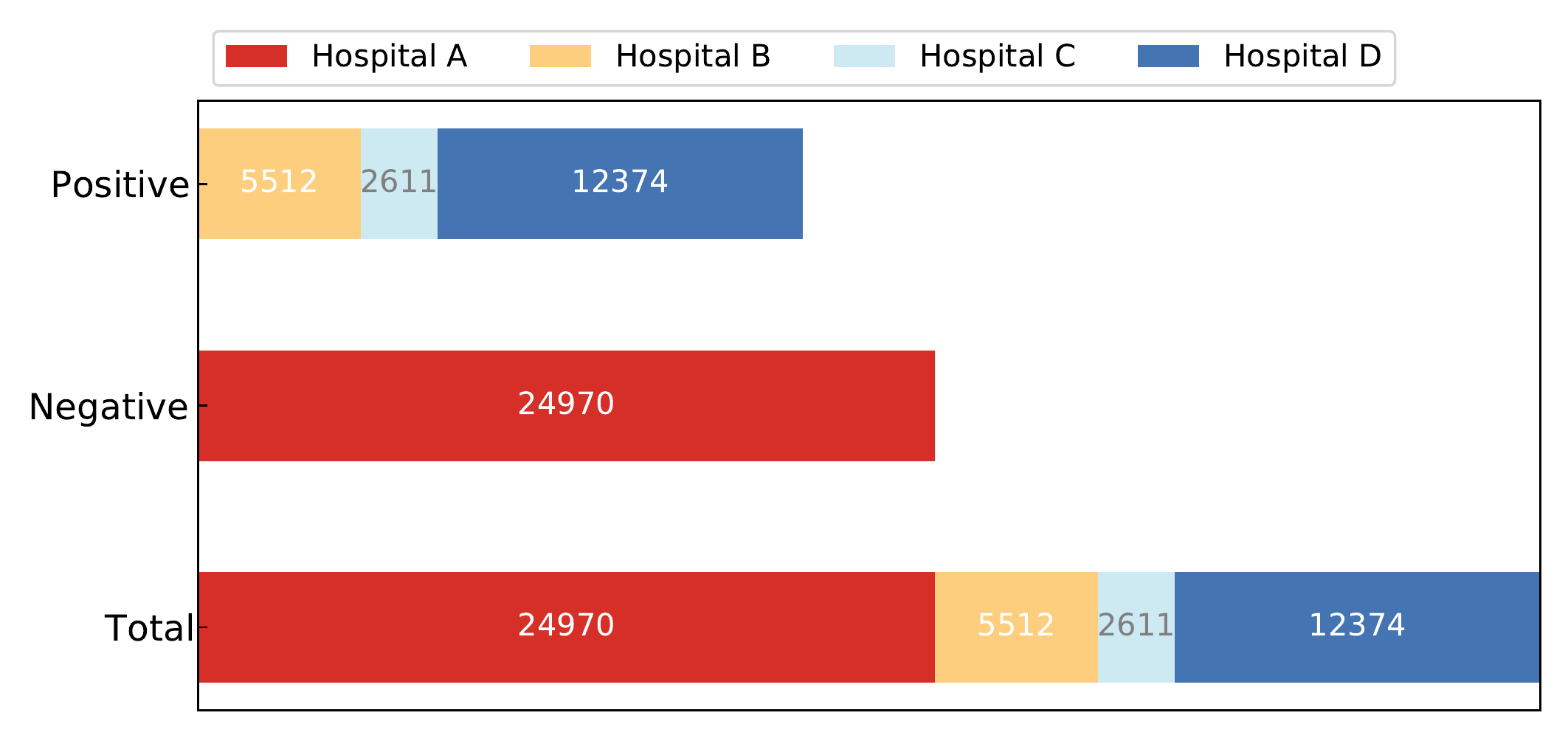} 
    }
    \caption{Class distribution of the collected CT data. The numbers in the sub-figures (a) and (b) represent the counts for the patient-level statistics and image-level statistics, respectively. The data collected from several clinical centres can result in great challenges in learning discriminative features from those class-imbalanced centres.}
    \label{fig:cls_dataset}
\end{center}
\end{figure}

We collected CT data from four different local hospitals in China and removed the personal information to ensure data privacy. The information of our collected data is summarised in Table \ref{fig:cls_dataset}. In total, there were 380 CT volumes of the patients who tested COVID-19 positive (reverse transcription polymerase chain reaction test confirmed) and 424 COVID-19 negative CT volumes. For a fair comparison, we trained the model on the cross-centre datasets collected from hospital A, B, C, and D. For an unbiased independent testing, CC-CCII data \cite{zhang2020ccii}, a publicly available dataset, which contained 2,034 CT volumes with 130,511 images, was adopted to verify the effectiveness of the trained models.

\subsubsection{Data Standardisation, Pre-Processing and Augmentation}
Following the protocol described in \cite{zhang2020ccii}, we used the U-Net segmentation network \cite{ronneberger2015unet} to segment the CT images. Then, we randomly cropped a rectangular region whose aspect ratio was randomly sampled in $[3/4, 4/3]$, the area was randomly sampled in $[90\%, 100\%]$, and the region was then resized into $224 \times 224$. Meanwhile, we randomly flipped the input volumes horizontally with 0.5 probability. The input data would be a set of CT volumes, which were composed of consecutive CT image slices.

\subsubsection{Quantitative Results}
We compared our proposed classification model with several state-of-the-art COVID-19 CT classification models \cite{he2016resnet, wang2020covid, li2020artificial, ouyang2020dual}. Table \ref{tb:sota} summarises the experimental results of COVID-19 classification on the CC-CCII data. For image-level annotations, ResNet-50 \cite{he2016resnet} and COVID-Net \cite{wang2020covid} simply treated patient-level labels as the image labels. Different from methods proposed by \cite{he2016resnet, wang2020covid, li2020artificial}, VBNet \cite{ouyang2020dual} utilised the 3D residual convolutional neural network to train with patient-level annotations on the whole CT volumes rather than single slices. Besides, COVNet \cite{li2020artificial} extracted prediction scores from each slice in the CT volumes with ResNet and aggregated the prediction scores via a max-pooling operator to get the patient-level probability. 

In Table \ref{tb:sota}, we can find that our method achieved the best performance among these SOTA methods. In particular, our method obtained a better performance by 7.2\% on AUC compared to VB-Net \cite{ouyang2020dual} on the patient-level indicating that our method can be applicable for the real-world scenario. This also verified the benefit of modelling section information in the CT volumes via our proposed SIM, which we believe is also vital to the improvement of the classification performance. Besides, our method significantly outperformed other methods by at least 40\% with respect to the specificity while maintaining high sensitivity, which is also a crucial indication for diagnosing COVID-19. In addition, models trained on patient-level annotations could achieve better performance compared to those trained on image-level labels. This is because the noise in the image labels could have a negative impact during the training, which might degrade the representation ability of the model. According to \cite{geirhos2018imagenet}, models trained on images may rely on learning the textures of images that were highly discriminative among multiple centres. Therefore, these trained models might be overfitted and biased to the texture features of the images collected from different centres, which could explain the phenomenon that these methods (i.e., \cite{he2016resnet, wang2020covid}) were poorly performed on the unseen centres.

In another aspect, for CT volumes, the sequential ordering of CT image slices is also informative. COVID-Net \cite{li2020artificial} took the most discriminative slice as the representation of the whole CT volume, which ignored the encoding of adjacent slices. This would enforce the model only detect the most discriminative slice, leading to the bias towards positive cases, which could impede the detecting of negative cases that resulted in a low specificity. On the contrary, VBNet proposed by Ouyang et al. \cite{ouyang2020dual} preserved the sequential information by training on the whole CT volumes. In contrast, we partitioned the CT volume into several sections in order to preserve the sequential information to some extent. Besides, VB-Net was trained with stronger supervision that it utilised additional masks for its supervised training. For our method, we only used patient-level annotations that were much more efficient. More importantly, our method achieved better performance on both AUC and accuracy compared to VBNet \cite{ouyang2020dual} and COVNet \cite{li2020artificial}.

\begin{table*}
\begin{center}
\resizebox{1.0\linewidth}{!}{%
\begin{tabular}{c|c|c|c|c|c|c}
\hline
\textbf{Annotation}            & \textbf{Method}                      & \textbf{Patient Acc. (\%)}  & \textbf{Precision (\%)} & \textbf{Sensitivity (\%)} & \textbf{Specificity (\%)} & \textbf{AUC (\%)} \\ \hline
\multirow{5}{*}{Patient-level} & ResNet-50 \cite{he2016resnet}        & 53.44                      & 64.45                   & 63.03                     & 35.71             & 53.24       \\
                               & COVID-Net \cite{wang2020covid}       & 57.13                      & 62.53                   & 84.70                     & 6.16             & 49.58      \\
                               & COVNet \cite{li2020artificial}       & 69.96                      & 70.20                   & \textbf{93.33}            & 26.75             & 81.61     \\
                               & VB-Net \cite{ouyang2020dual}         & 76.11                      & 75.84                   & 92.73                     & 45.38             & 88.34      \\
                               & Ours                                 & \textbf{89.97}             & \textbf{92.99}          & 91.44                     & \textbf{87.25}    & \textbf{95.53}      \\ \hline
\multirow{4}{*}{Image-level}   & ResNet-50 \cite{he2016resnet}        & 52.56                      & 61.60                   & 71.27                     & 18.06     &   50.19              \\
                               & COVID-Net \cite{wang2020covid}       & 60.03                      & 64.81                   & \textbf{83.91}            & 15.98     &   58.39            \\
                               & COVNet \cite{li2020artificial}       & 75.55                      & 79.90                   & 83.24                     & 61.37     &   79.48            \\
                               & Ours                                 & \textbf{80.41}             & \textbf{88.56}          & 80.15                     & \textbf{80.89}  &  \textbf{86.06}      \\ \hline
\end{tabular}
}
\end{center}
\caption{Comparison results of our method vs. state-of-the-art methods performed on the CC-CCII dataset.}
\label{tb:sota}
\end{table*}

\begin{figure}
\begin{center}
    \subfigure[Patient-level Annotation]{
        \includegraphics[width=0.46\linewidth]{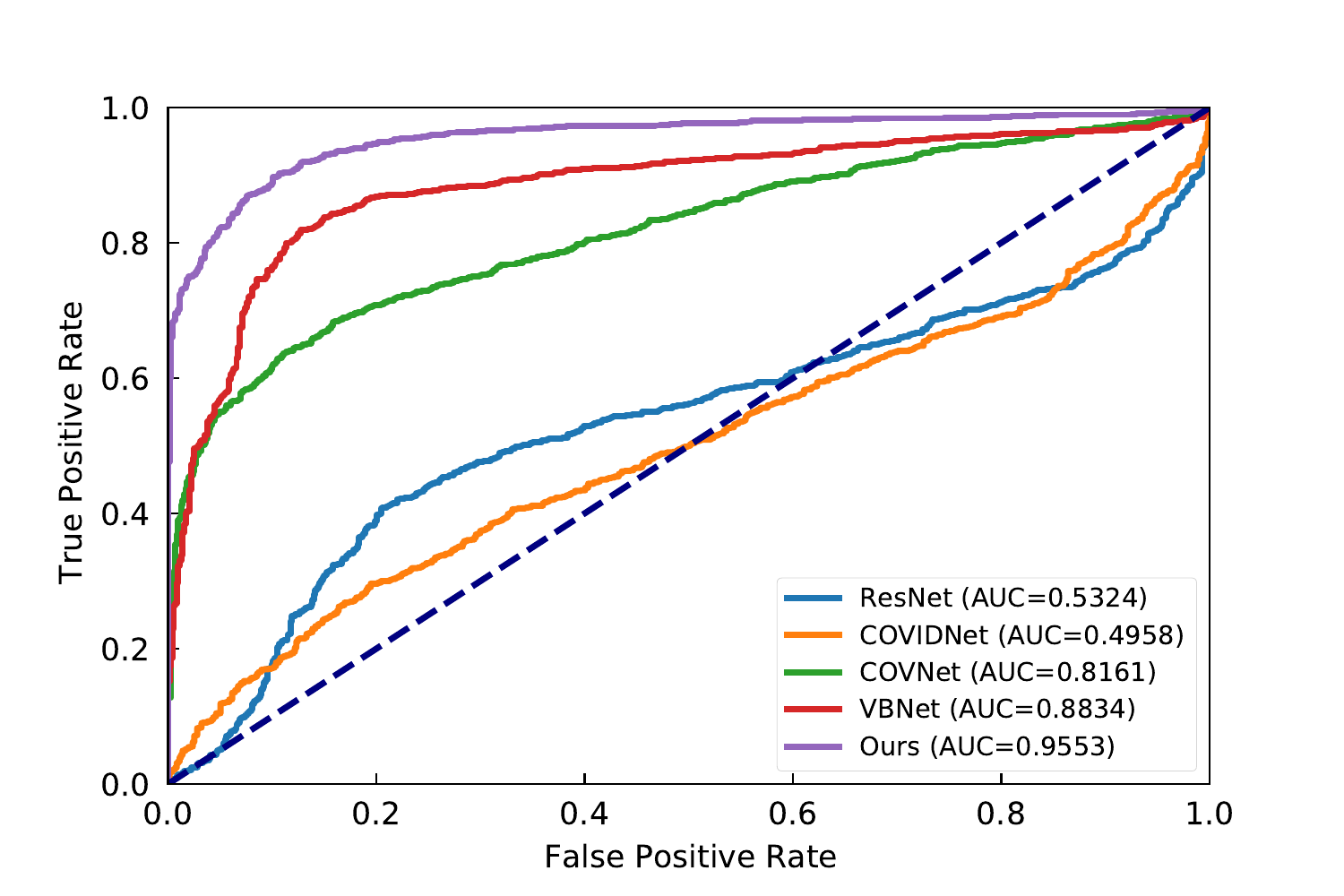} 
    }
    \quad
    \subfigure[Image-level Annotation]{
        \includegraphics[width=0.46\linewidth]{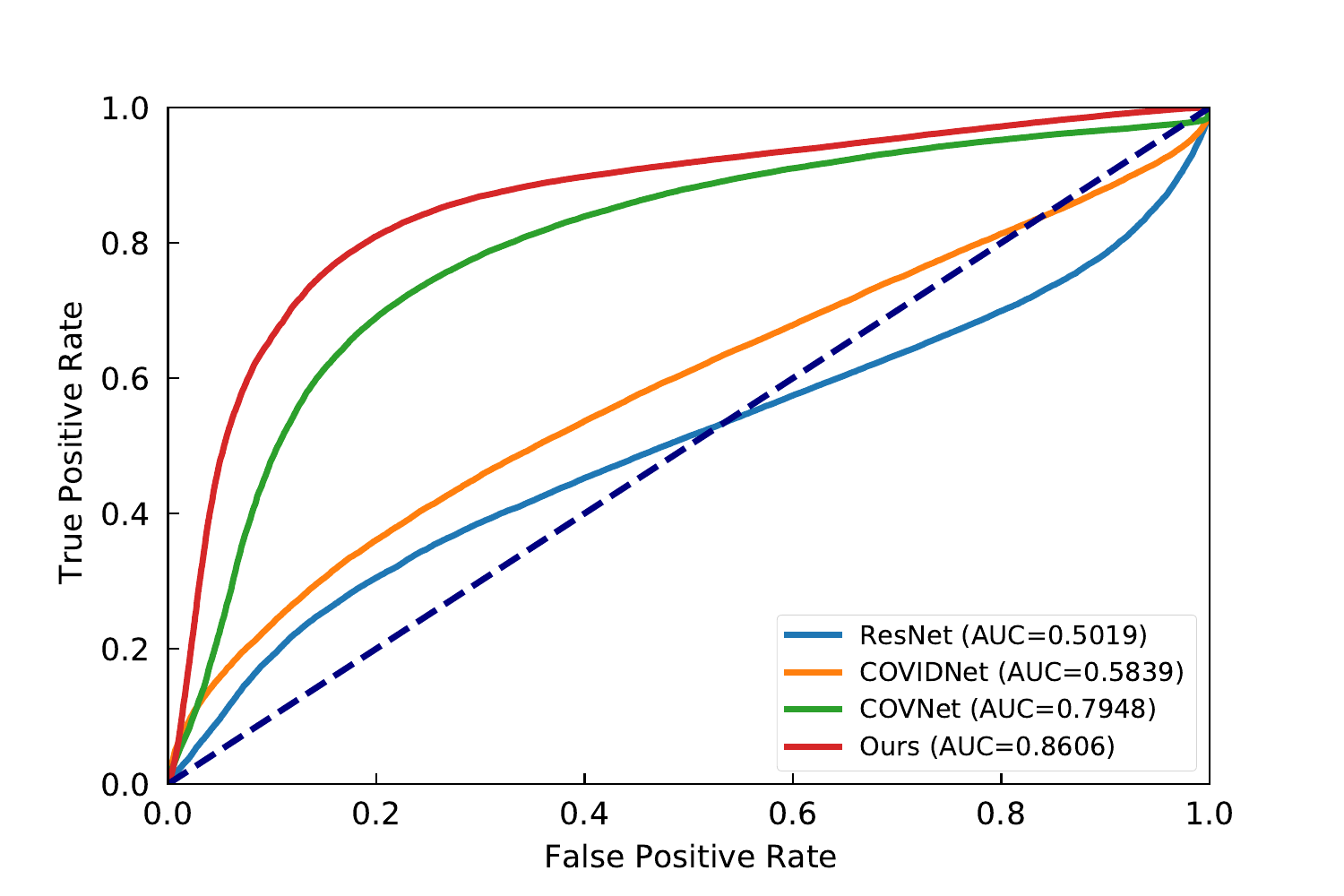} 
    }
    \caption{The Receiver Operating Characteristic (ROC) curves of different compared methods.}
    \label{fig:roc}
\end{center}
\end{figure}

\begin{figure}
\begin{center}
    \subfigure[Patient-level Annotation]{
        \includegraphics[width=0.46\linewidth]{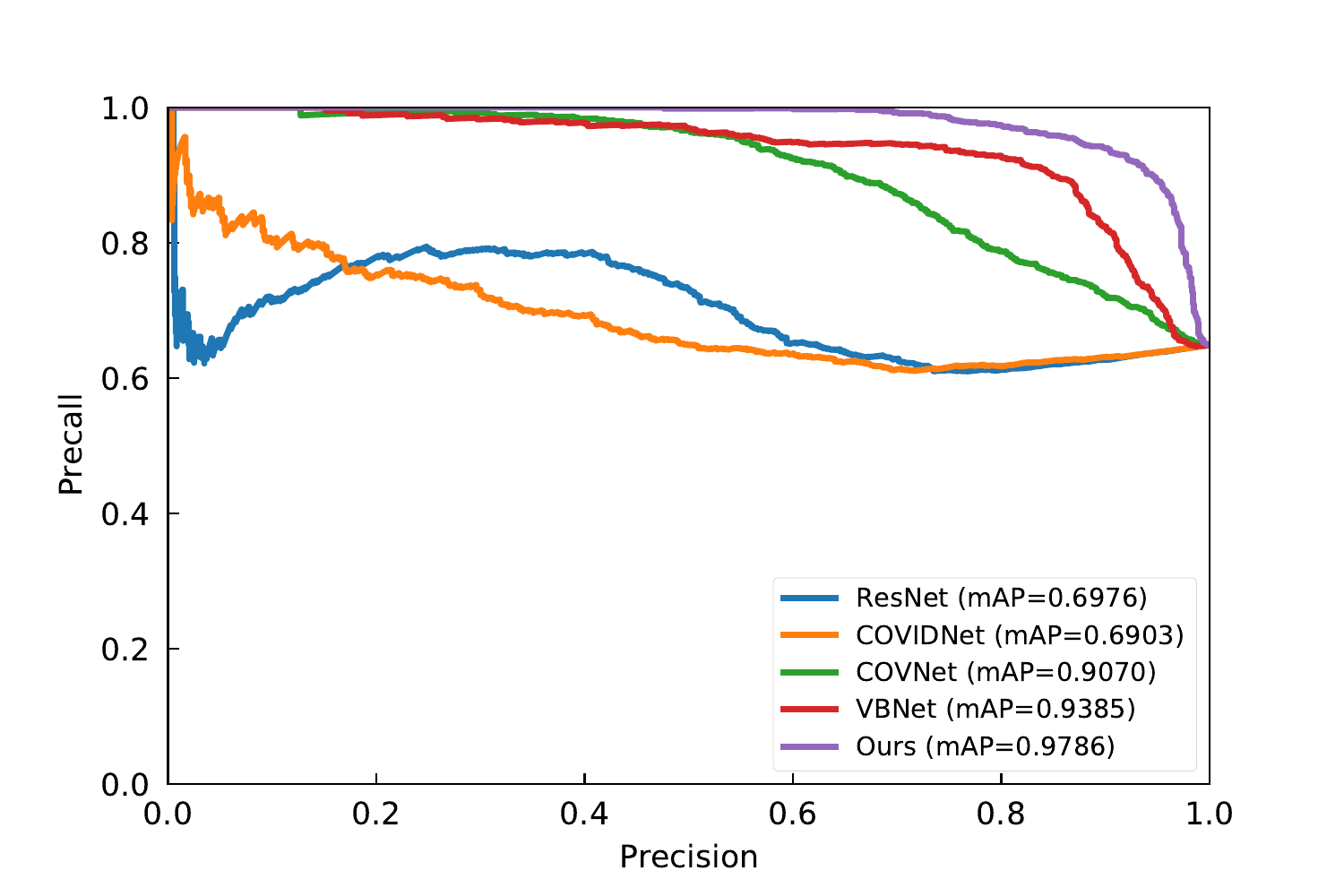} 
    }
    \quad
    \subfigure[Image-level Annotation]{
        \includegraphics[width=0.46\linewidth]{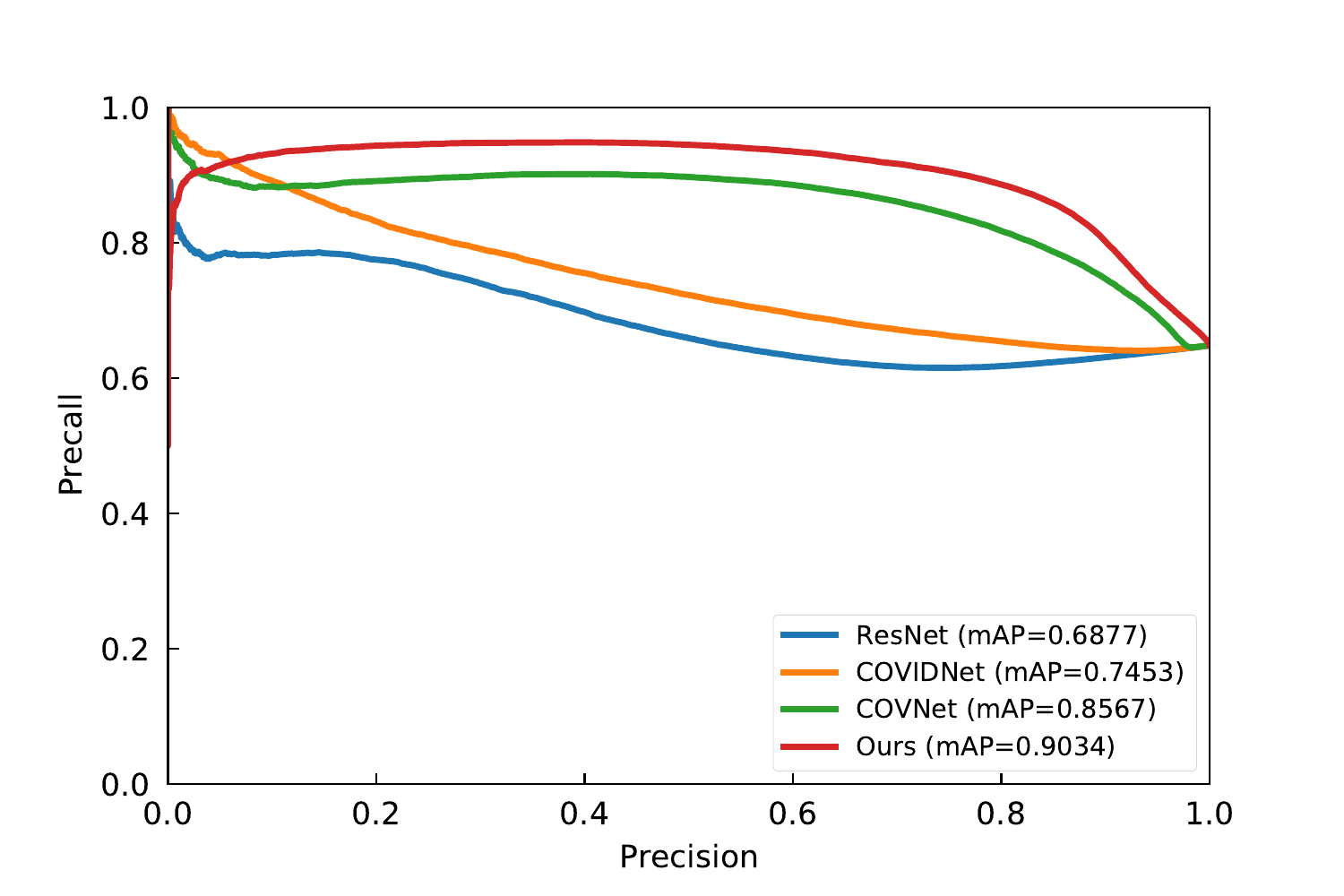} 
    }
    \caption{The Precision-Recall (PR) curves of different compared methods.}
    \label{fig:pr}
\end{center}
\end{figure}

In addition, we also provided the Precision-Recall (PR) and Receiver Operating Characteristic (ROC) curves to compare different methods on patient-level annotations and image-level annotations (Figure \ref{fig:pr} and \ref{fig:roc}). From the figure, we can observe that models trained on image-level annotations (e.g., ResNet \cite{he2016resnet} and COVID-Net \cite{wang2020covid}) were poorly performed since their AUCs were close to 50\% which indicated a random guess. In contrast, models trained on patient-level was more reliable since their AUCs were greater than 50\%. In particular, we found that overall our proposed method remained the best-performed algorithm with an AUC of 95.53\% at the patient-level and 86.06\% at the image-level. These results verified our assumption that for mild COVID-19 cases, most of the image slices are disease-free.

\subsubsection{Qualitative Results}

\begin{figure}
\begin{center}

    \includegraphics[width=\linewidth]{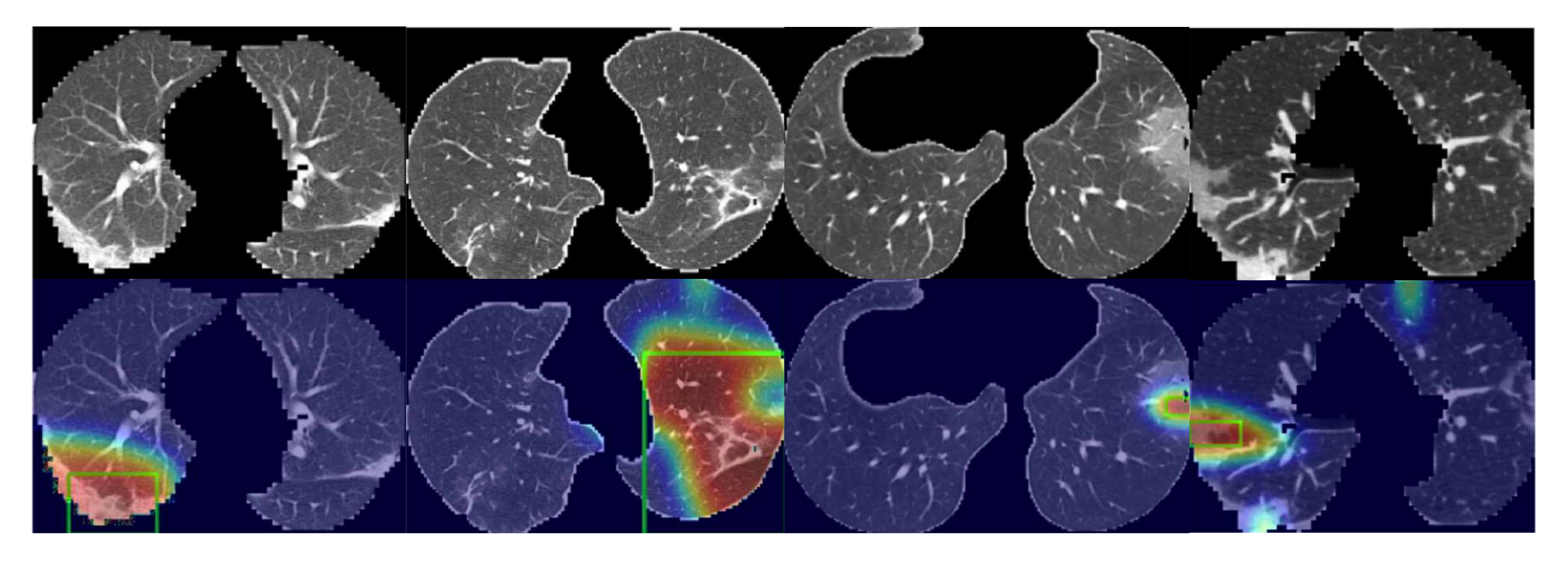} 
    \caption{Examples of the CAMs $A^{conv}$ generated by our proposed EDM for classifying COVID-19 positive patients. The first row contains the original CT-scan image slices, and the second row illustrates the heatmaps of CAMs $A^{conv}$ with bounding boxes confined to the infected areas.}
    \label{fig:cls_cam}
\end{center}
\end{figure}

In order to make the prediction to be more explainable, we used the trained model to visualise the CAMs and bounding boxes generated by our EDM as described above. Figure \ref{fig:cls_cam} shows the visualisation results of the derived CAMs (i.e., $A^{conv}$). In this figure, we can clearly observe that our method tended to pay more attention to the discriminative part of the images so as to make the predictions. For example, in the first column, the lower left part of the lung was seriously infected and had a large area of lesions. Therefore, our method would make the predictions that the image was classified as COVID-19 positive, demonstrating the capability of our XAI model to make explainable predictions.

In addition, based on the results of the derived CAMs, we also extracted the lesion bounding boxes from the CAMs. It can be found that our method was capable of yielding accurate bounding boxes from the salient part of the CAMs, as illustrated in Figure \ref{fig:cls_cam}, which further confirmed that our XAI method was applicable to be an auxiliary diagnosis tool for the clinicians.

\begin{figure*}[!ht]
\centering
\includegraphics[width=\textwidth]{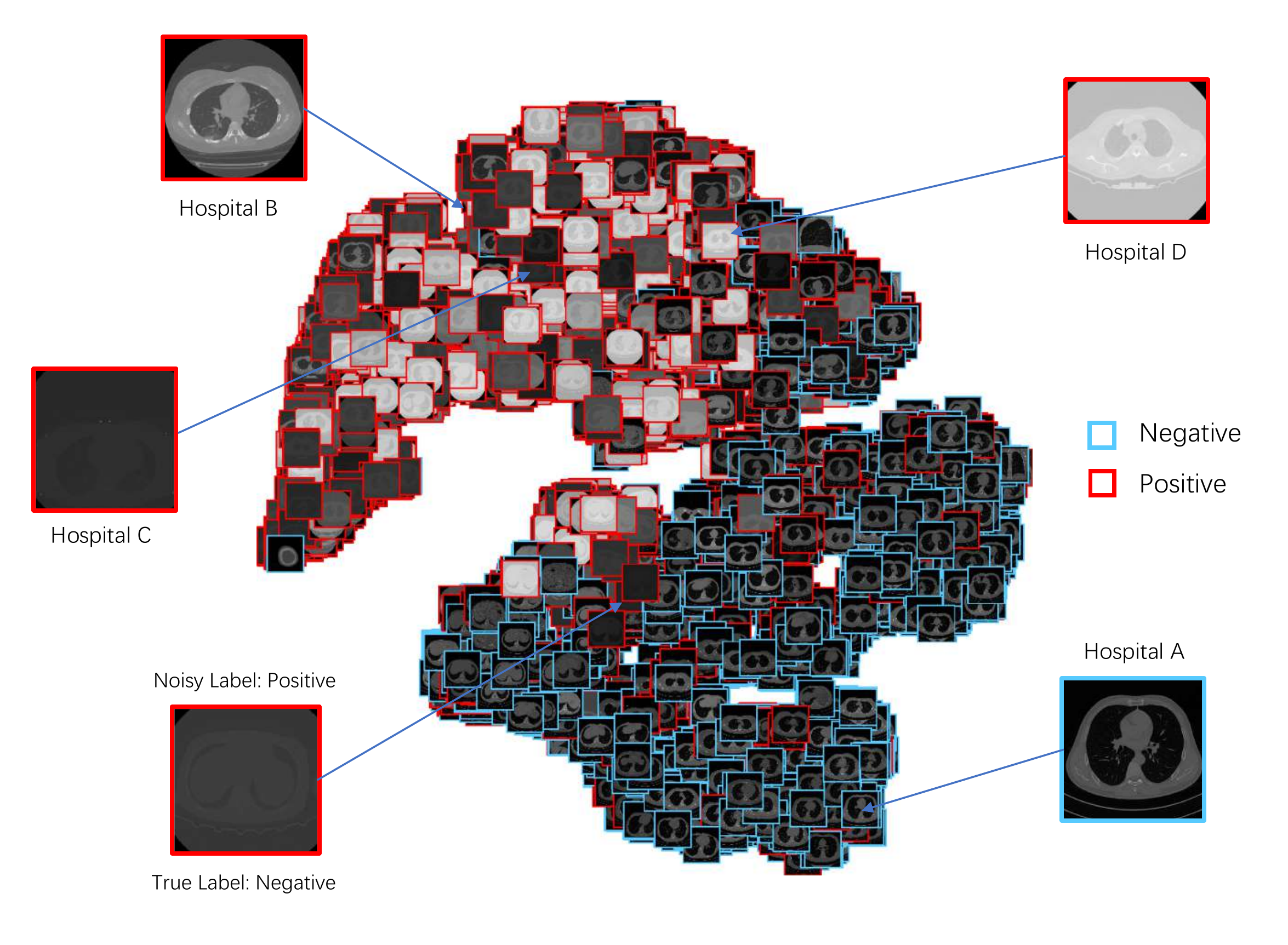}
\caption{T-SNE visualisation \cite{maaten2008tsne} of the learnt features from CT images. Original images are sampled from four different hospitals and represented in the figure. Besides, a falsely annotated image is drawn from the negative cluster.}
\label{fig:t_sne}
\end{figure*}

To further illustrate the learnt features from our proposed method, we extracted the feature from the backbone network of our architecture, and used T-SNE \cite{maaten2008tsne} visualisation technique to transform the features extracted from the backbone network of our proposed model, and visualised the distribution of the classified images as shown in Figure \ref{fig:t_sne}. In this figure, we can find the distinctive visual characteristics of the CT images from different hospitals (i.e., Hospital A, B, C, and D). Besides, it can be observed that the COVID-19 positive images were mostly clustered together, and negative images were mainly distributed in another cluster. More interestingly, in the cluster of the negative images, we can find several positive images in this cluster since these images were scanned from patients who were tested COVID-19 positive. In our intuition, we assume that for some mild cases, lesions were not presented in all of the CT slices. Therefore, there were indeed disease-free CT slices that could be falsely labelled as COVID-19 positive, which verified our assumption.

\begin{figure*}
\centering
\includegraphics[width=\textwidth]{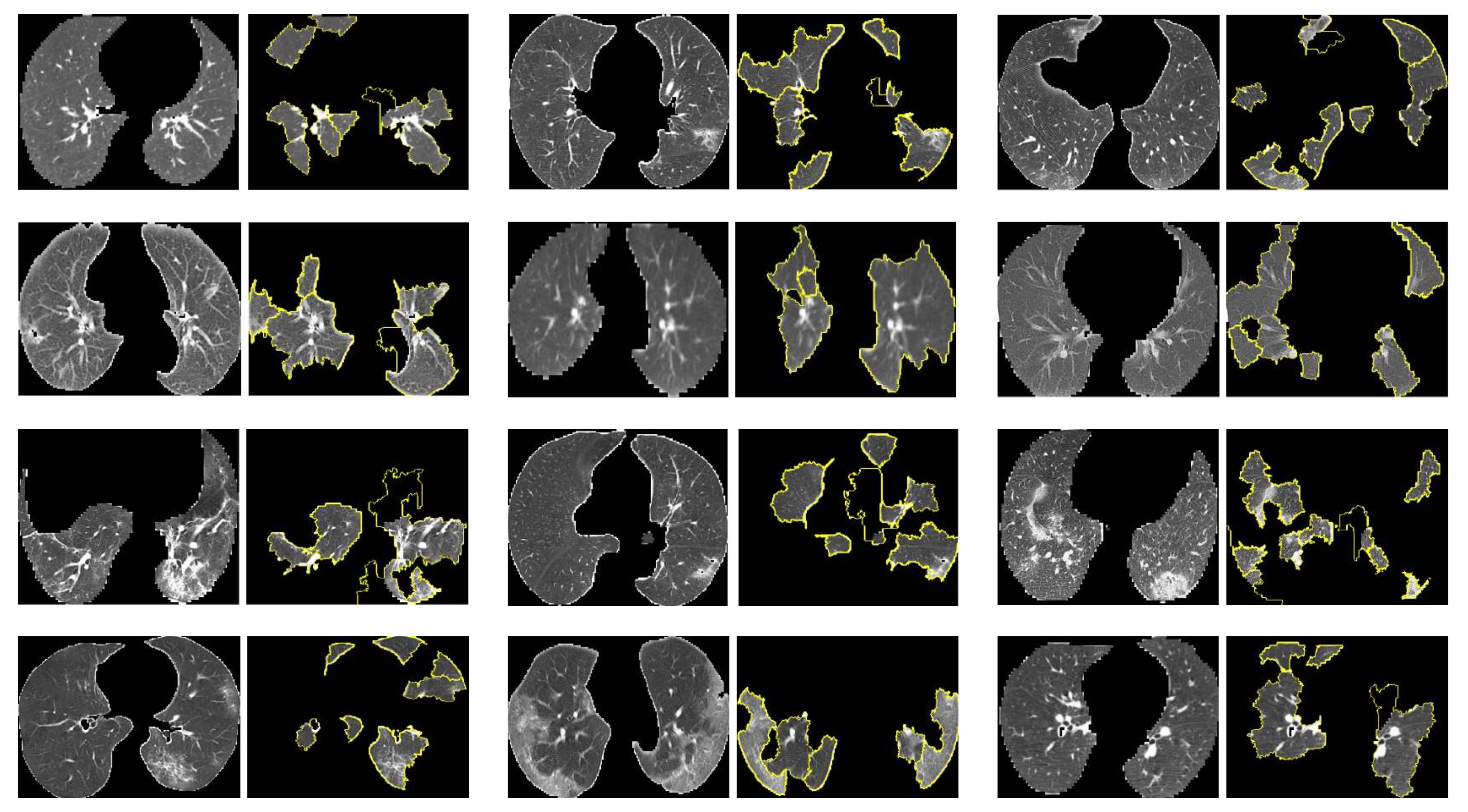}
\caption{Visualisation of the super-pixels that are positively contributed to the predictions via the LIME method \cite{ribeiro2016lime}.}
\label{fig:lime}
\end{figure*}

Additionally, in order to explain each individual prediction, we adopted the LIME method \cite{ribeiro2016lime} to investigate the contribution of each pixel for the prediction. Instead of using the individual pixel, we divided an image into super-pixels, which were composed of interconnected pixels with similar imaging patterns. Figure \ref{fig:lime} shows the explanations via LIME for COVID-19 positive images. In each pair of images, we visualised the super-pixels that contributed to the COVID-19 positive prediction results. We can observe that the lesion parts would explain for the positive prediction, which is reasonable to our deep learning model.

\begin{figure*}[!ht]
\centering
\includegraphics[width=\textwidth]{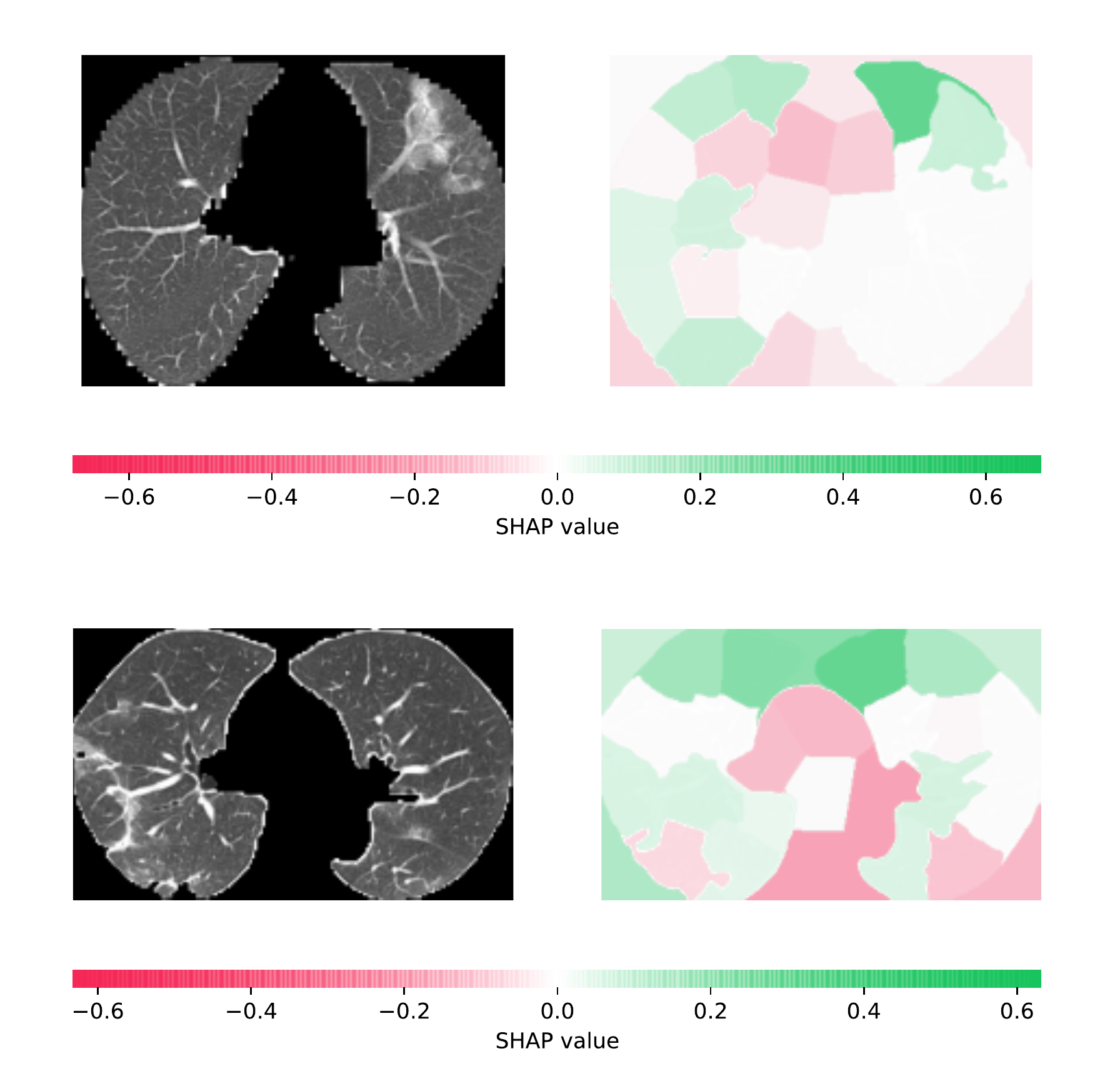}
\caption{The SHAP values for different super-pixels of the sampled images. We computed the SHAP values through the Kernel SHAP method \cite{lundberg2017shap}. The super-pixel with positive SHAP value indicates the positive impact to the positive prediction, while the negative value means that the super-pixel contributes to the negative prediction.}
\label{fig:shap}
\end{figure*}

However, the LIME method could only quantitatively estimate the importance according to how close the combination of super-pixels was to the original instance. It discarded the global view of the individual feature contributed by the super-pixels. To overcome this drawback, we further leveraged Kernel SHapley Additive exPlanations (Kernel SHAP) method \cite{lundberg2017shap} to estimate the contribution of each super-pixel quantitatively by the SHAP value. Samples explained by the Kernel SHAP are demonstrated in Figure \ref{fig:shap}. We can observe that the super-pixels contained lesion areas positively contributed to the positive prediction, while those super-pixels related to the backgrounds or disease-free areas would reflect the contribution to negative prediction.

\subsection{Showcase II: Segmentation for Hydrocephalus} 

\subsubsection{Datasets}
The studied cohort included 20 normal elderly people, 20 patients with cerebral atrophy, 64 patients with normal pressure hydrocephalus, and 51 patients with acquired hydrocephalus (caused by subarachnoid haemorrhage, brain trauma or brain tumour). CT scans of the head were performed using two CT instruments, one of which was the SOMATOM Definition Flash from Siemens, Germany, and the other was the SOMATOM Emotion 16 from Siemens, Germany. Secondly, MRI examinations were conducted using a 1.5T MR scanner(Avanto, Siemens, Erlangen, Germany) and a 3.0T MRI scanner(Prisma, Siemens, Erlangen, Germany). The slice thickness of the CT images includes: 0.5mm, 1.0mm, 1.5mm, 2.0mm, 4.8mm, 5.0mm. The slice thickness of the MRI images includes: 1.0mm, 7.8mm, 8.0mm. For experiments, we randomly split the thick-slice and thin-slice images into training, validation and testing sets. The details of the dataset are summarised in Table \ref{tb:seg_dataset}.

\begin{table}[]
\begin{center}
\resizebox{1.0\linewidth}{!}{%
\begin{tabular}{l|c|c|c|c|cc}
\hline
\multirow{2}{*}{Modality} & \multicolumn{2}{c|}{Training Set} & \multicolumn{2}{c|}{Validation Set} & \multicolumn{2}{c}{Test Set}                  \\ \cline{2-7} 
                          & Thick-slice      & Thin-Slice     & Thick-slice       & Thin-Slice      & \multicolumn{1}{c|}{Thick-slice} & Thin-Slice \\ \hline
MRI                       & 810              & 1,303          & 203               & 326             & \multicolumn{1}{c|}{189}         & 982        \\
CT                        & 2,088            & 2,076          & 523               & 519             & \multicolumn{1}{c|}{309}         & 492        \\ \hline
\end{tabular}
}
\end{center}
\caption{The number of thick-slice and thin-slice images used in our study.}
\label{tb:seg_dataset}
\end{table}

\subsubsection{Data Standardisation, Pre-Processing and Augmentation}
For the pre-processing of these data, we normalised images using the $z$-score normalisation scheme, which was done by subtracting its mean then divided by its standard deviation. For anomaly pixels, we clipped them within the range of 1-quantile and 99-quantile. For data augmentation, we resized the images using a bicubic interpolation method and resized masks with the nearest interpolation. Then we flipped the images horizontally with 0.5 probability, and scaled the hue, saturation, and brightness with coefficients uniformly drawn from $[0.8, 1.2]$. 

\subsubsection{Quantitative Results}
Table \ref{tb:Dice_seg} shows the segmentation performance of various compared models on different modalities. All of the models were trained on the thick-slice images with annotations and the unlabelled thin-slice images. We can observe that our proposed method outperformed all of the compared state-of-the-art methods by a large margin on the mixed datasets (i.e., the mixture of thick-slice and thin-slice images) with at least 4.4\% of the Dice scores. It is of note that all three models achieved similar segmentation performance on thick-slice images. However, our proposed method gained a significant improvement on the thin-slice images for both MRI and CT scans. The primary reason is that our model could diminish the uncertainty of these thin-slice images while achieving the distribution alignment between thick-slice and thin-slice images, which could enhance the representation and generalisation capabilities of our model. Besides, we also investigated the effectiveness of $\mathcal{L}_\mathcal{S}$ and $\mathcal{L}_\mathcal{T}$ in Table \ref{tb:seg_ablation}. In the table, we can find that when only training on thick-slice images, the model performed perfectly on thick-slice images while performing poorly on thin-slice images, since the distribution of these two kinds of slices could vary. Moreover, the performance of the models trained only on unlabelled thin-slice images degraded sharply because of the lack of annotations to guide the segmentation. In the Exp.3 as shown in Table \ref{tb:seg_ablation}, our model could gain significant improvement on the thin-slice images while preserving good performance on the thick-slice images, which demonstrated that our trained model was applicable for both types of images for both CT and MRI modalities.

\begin{table}[]
\begin{center}
\begin{tabular}{l|c|c|c|ccc}
\hline
\multirow{2}{*}{Method} & \multicolumn{3}{c|}{MRI}                                                                & \multicolumn{3}{c}{CT}                                                                 \\ \cline{2-7} 
                        & Thick                       & Thin                        & Mixed                       & \multicolumn{1}{c|}{Thick}  & \multicolumn{1}{c|}{Thin}   & Mixed                      \\ \hline
U-Net \cite{ronneberger2015unet}                   & 0.9226                      & 0.7665                      & 0.8353                      & \multicolumn{1}{c|}{0.9351} & \multicolumn{1}{c|}{0.7987} & 0.8513                     \\
U-Net++ \cite{zhou2018unet++}                & \multicolumn{1}{l|}{0.9159} & \multicolumn{1}{l|}{0.8495} & \multicolumn{1}{l|}{0.8602} & \multicolumn{1}{l|}{\textbf{0.9421}} & \multicolumn{1}{l|}{0.7797} & \multicolumn{1}{l}{0.8424} \\
Ours                    & \textbf{0.9323}                  & \textbf{0.9056}                      & \textbf{0.9099}                      & \multicolumn{1}{c|}{0.9365} & \multicolumn{1}{c|}{\textbf{0.8697}} & \textbf{0.8954}                     \\ \hline
\end{tabular}
\end{center}
\caption{Comparison results (Dice scores) of our method vs. other state-of-the-art methods. Mixed represents the test set containing both thick-slice and thin-slice images.}
\label{tb:Dice_seg}
\end{table}

\begin{table}[]
\begin{center}
\begin{tabular}{l|c|c|c|c|c|ccc}
\hline
\multirow{2}{*}{\textbf{Exp.}} & \multirow{2}{*}{$\mathcal{L}_\mathcal{S}$} & \multirow{2}{*}{$\mathcal{L}_\mathcal{T}$} & \multicolumn{3}{c|}{\textbf{MRI}}                   & \multicolumn{3}{c}{\textbf{CT}}                                                               \\ \cline{4-9} 
                               &                             &                             & Thick           & Thin            & Mixed           & \multicolumn{1}{c|}{Thick}           & \multicolumn{1}{c|}{Thin}            & Mixed           \\ \hline
1                              & $\surd$                     &                             & \textbf{0.9390} & 0.8199          & 0.8391          & \multicolumn{1}{c|}{\textbf{0.9438}} & \multicolumn{1}{c|}{0.8345}          & 0.8767          \\
2                              &                             & $\surd$                     & 0.0034          & 0.0108          & 0.0110          & \multicolumn{1}{c|}{0.0109}          & \multicolumn{1}{c|}{0.0006}          & 0.0069          \\
3                              & $\surd$                     & $\surd$                     & 0.9323          & \textbf{0.9056} & \textbf{0.9099} & \multicolumn{1}{c|}{0.9365}          & \multicolumn{1}{c|}{\textbf{0.8697}} & \textbf{0.8954} \\ \hline
\end{tabular}
\end{center}
\caption{Dice scores comparison for verifying the effectiveness of each loss term. Mixed represents the test set containing both thick-slice and thin-slice images.}
\label{tb:seg_ablation}
\end{table}

\begin{figure*}[!ht]
\centering
\includegraphics[width=\textwidth]{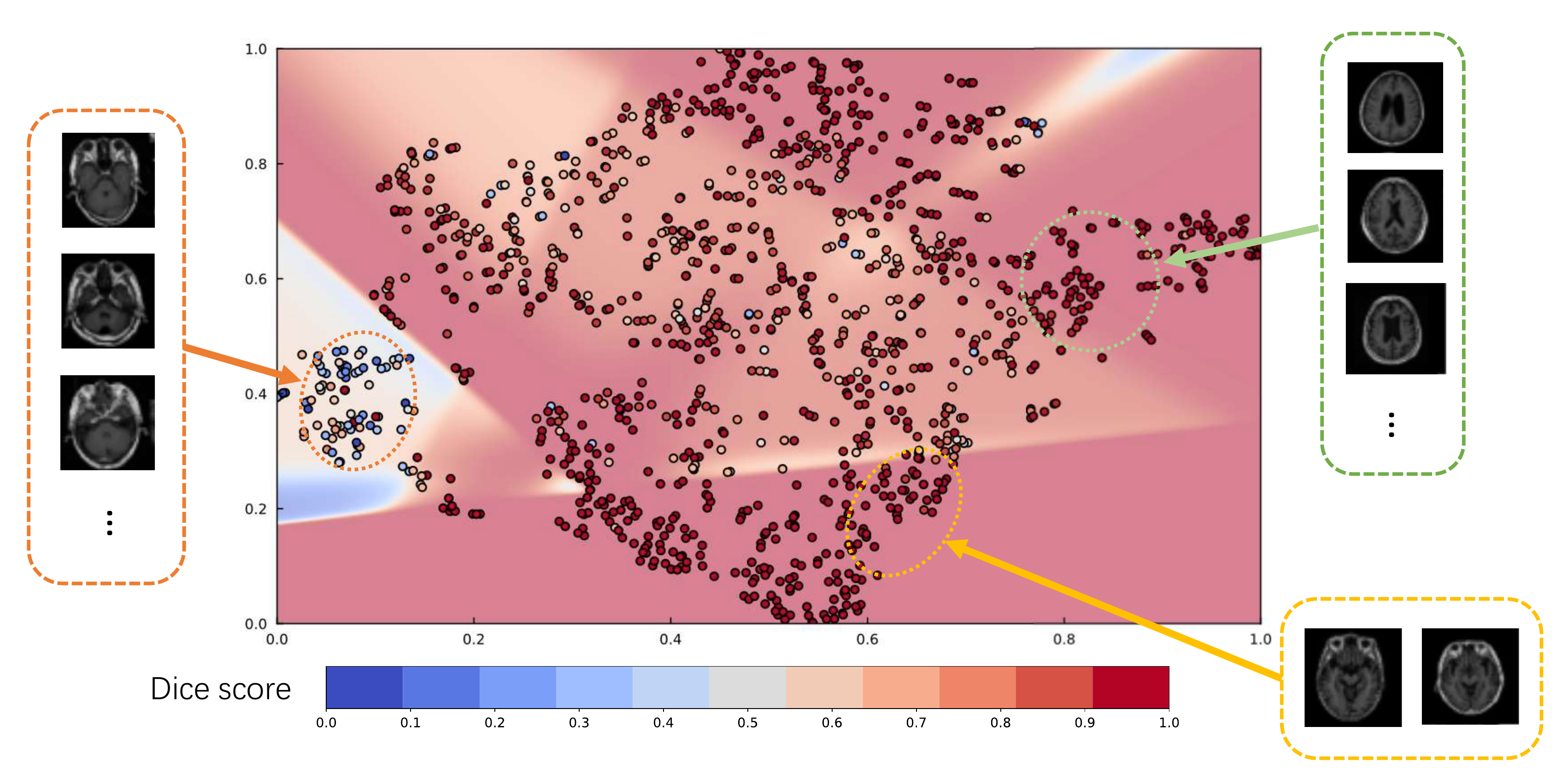}
\caption{The visualisation of the Dice scores of the projected images. The plane was computed by smoothing the Dice scores. It is of note that images sharing similar characteristics were clustered together. On the left-hand side and right-hand side, samples from different regions of the plane are presented.}
\label{fig:ious}
\end{figure*}

Besides, in order to interpret the black-box segmentation model, we extracted the lowest bottom features and projected them into a 2D latent space using the PCA technique. We then computed the Dice score for each sample and visualised it in Figure \ref{fig:ious}. In this figure, we can observe that slices sampled from the orange circle all contained a small region of ventricle where the model could not perform well. However, images from the green and yellow circle had multiple ventricles, which took a large proportion of the images. Therefore, these images could be well-predicted by our model.

\subsubsection{Qualitative Results}
\begin{sidewaysfigure}
\begin{center}

    \includegraphics[width=\linewidth]{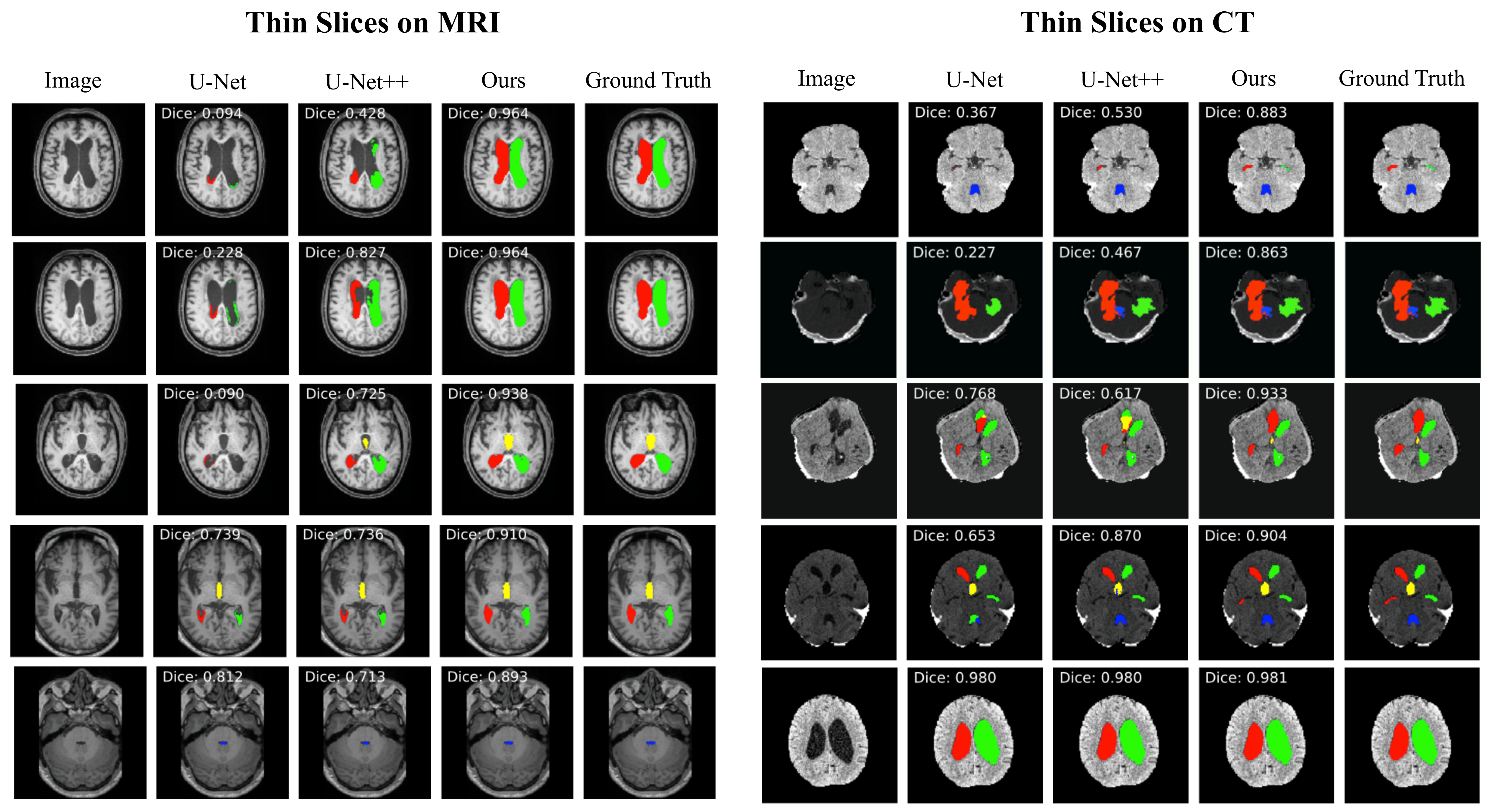} 
    \caption{The visualisation of the 3D brain ventricles segmentation results using different compared models. The right lateral ventricle is coloured in red; the left lateral ventricle is coloured in green; the yellow coloured region represents the third ventricle; and the blue region represents the fourth ventricle.}
    \label{fig:seg_results}
\end{center}
\end{sidewaysfigure}

To qualitatively examine the performance of our model and other state-of-the-art models, we presented some visualisation results of the CT and MRI images with thin-slices in Figure \ref{fig:seg_results}, and computed the Dice scores for each segmentation result. For MRI images, our model and U-Net++ \cite{zhou2018unet++} were able to segment four ventricles in the brain. In particular, our model could predict the third ventricle in the brain more completely compared to the prediction generated by the U-Net++ \cite{zhou2018unet++} due to the informative feature representation by the pre-trained encoder. However, for CT images, the performance varied among different models. The primary reason is that original CT volumes contained the skull which could cause the brain to be visually unclear, after removing the skull, the contrast of the images could be largely distinct. More concretely, for those images with low contrast, (e.g., the row 1 and row 5 in Figure \ref{fig:seg_results}), all of the three compared methods were capable of predicting the left lateral and right lateral ventricles. However, for those images with high contrast (e.g., the row 2 and row 4 in Figure \ref{fig:seg_results}), our proposed method could predict most of the ventricle part in the brain while U-Net and U-Net++ failed.

\begin{figure}[!ht]
\begin{center}

    \includegraphics[width=\linewidth]{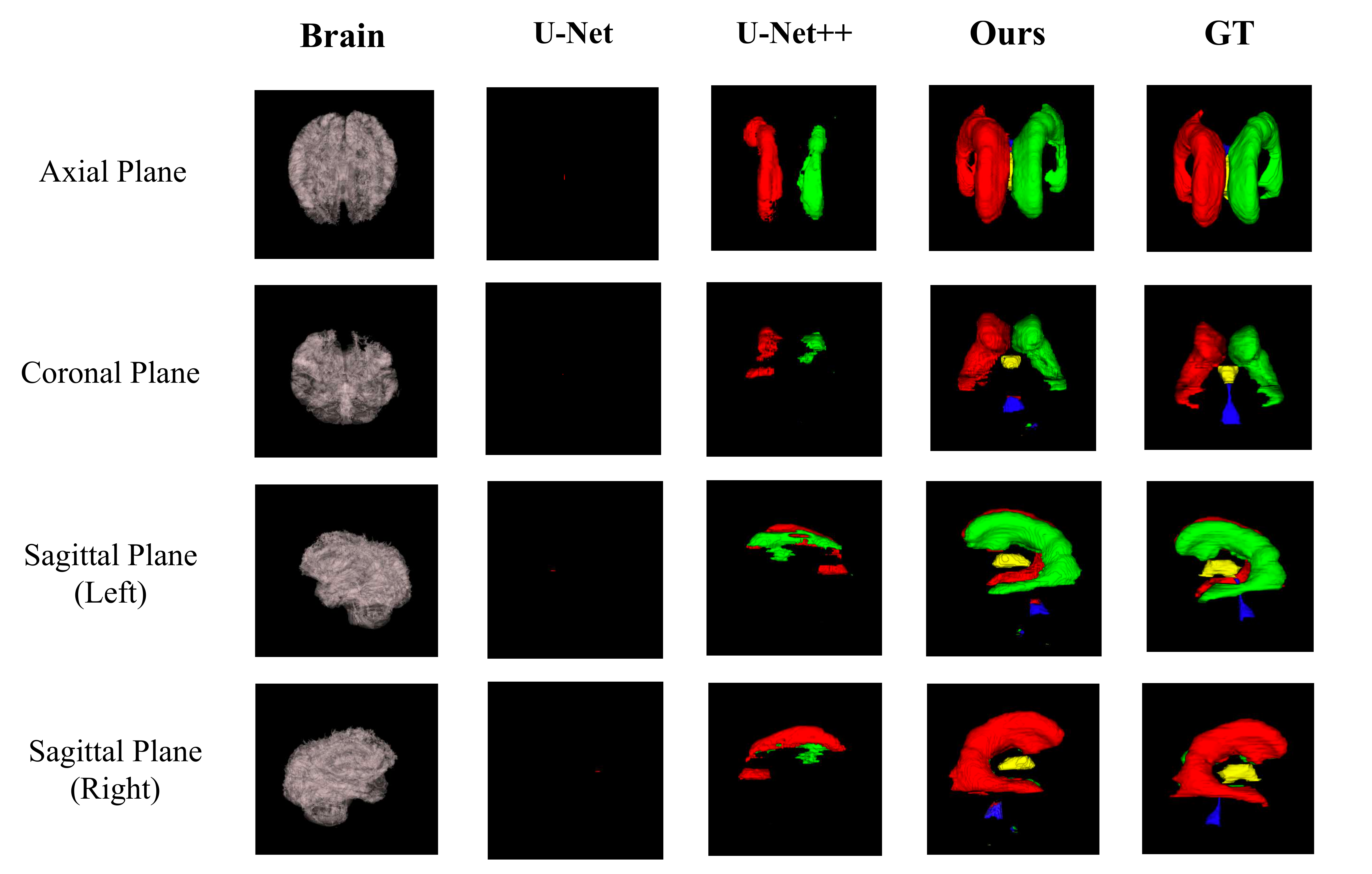} 
    \caption{Three-dimensional visualisation of the predictions on thin-slice MRI images for each ventricle segmented by different comparison models. The 3D segmentation results were visualised from the axial plane, the coronal plane, and the sagittal plane. Colouring scheme is consistent with Figure \ref{fig:seg_results}.}
    \label{fig:seg_3d_results}
\end{center}
\end{figure}

In addition, we used the segmentation results generated by compared models to reconstruct the 3D images of each ventricle. The example is illustrated in Figure \ref{fig:seg_3d_results}. We can observe that U-Net \cite{ronneberger2015unet} could hardly predict the ventricles on thin-slice images, while U-Net++ \cite{zhou2018unet++} was able to segment the left lateral and right lateral ventricles by taking advantage of dense connections of the intermediate feature maps. In contrast, our proposed method could not only predict the two ventricles mentioned above, but could also segment the third ventricle and the fourth ventricle well. One limitation of our model is that it could not predict the connection region between the third ventricle and fourth ventricle because the area is too small to be distinguished.

\subsection{Discussions}

In missions increasingly vital to human healthcare, AI is being deployed. Automated decisions should be explainable in order to create trust in AI and prevent an algorithm-based totalitarian society. This is not just a human right, for example, enshrined in the European GDPR, but an ultimate goal for algorithm developers who want to know if the necessary clinical characteristics are captured by the decision support systems. XAI should be possible to provide explanations in a systematic manner in order to make the explainability scalable. To construct a surrogate while-box model for the black-box model used to make a prediction, a typical solution is to use simpler, more intuitive decision algorithms. There is a chance, though, that the surrogate model is too complicated or too abstract for it to be truly understandable for humans. 

In this study, we have firstly provided a mini-review for XAI methods and their specific applications in medicine and digital healthcare that is followed by two example showcases that we have developed. From our two showcases, we have explored the classification model and segmentation model in terms of sensitivity (i.e., LIME \cite{ribeiro2016lime} and Kernel SHAP \cite{lundberg2017shap}) and decomposition (i.e., T-SNE \cite{maaten2008tsne} and CAMs). For LIME and Kernel SHAP methods, the individual sample can be analysed and interpreted with each super-pixel, which is useful for individual diagnosis. These methods can provide a straightforward view of how local explanations affect the final predictions. 

On the other hand, T-SNE provides us with an insight into the strength and weakness of our proposed models. For example, in Figure \ref{fig:ious}, the distribution of the decomposed image features has an association with the prediction performance, which indicates the weakness of the black-box segmentation models. Meanwhile, the distribution of decomposed image features also reveals the clustered characteristics of the raw inputs (Figure \ref{fig:t_sne}), which can help us to find the reason why a model would make such predictions.

In consequence, these methods can also be classified into two categories named as perceptive interpretability and mathematical interpretability. When visual evidence is not useful or erroneous, the mathematical evidence can be used as the complement for interpretability. Therefore, various methods should be applied simultaneously for the sake of providing reliable interpretability.

Nevertheless, a significant drawback of the current studies on XAI is that the interpretations are focused on the intuition of experts rather than from the demands of the end-users \cite{du2019techniques}. Current local explanations are typically provided in a feature-importance vector format, which is a full causal attribution and a low-level interpretation. This format would be satisfactory if the description viewers were the developers and analysts, since they could use the mathematical study of the distribution of features to debug the models. However, this type of XAI is less accommodating if the description receivers are lay-users of the AI. XAI can explain the complete judgement logic of the model, which includes a large amount of repetitive knowledge which can confuse the lay-users. The presentation of the XAI algorithms should be further improved to increase customer satisfaction.

The poor abstraction level of explanations is another drawback. For example, despite XAI derived heatmaps can indicate that individual pixels are important, there is normally no correlation computed between these significance regions to more abstract principles such as the anatomical or pathological regions shown in the images. More importantly, the explanations ought to be understood by humans to make sense of them and to grasp the understandable actions of the model. It is indeed desirable to provide meta-explanations that can integrate evidence from these low-level heatmaps to describe the behaviour of the model at a more abstract, more humanly understandable level. However, this level of understanding can be hard and erroneous. Previously proposed methods have recently been suggested to aggregate low-level explanations and measure the semantics of neural representations. Thus, a constructive topic for future study is the development of more advanced meta-explanations that leverages multimodal information fusion.

Because the audiences of XAI results are essentially human users, an important future research direction is the use of XAI in human-machine interaction; therefore, research studies in XAI need to explore human factors. A prerequisite for good human-machine interaction is to construct explanations for the right user focus, for instance, develop XAI to ask the correct questions in the proper manner, which is crucial in the clinical environment. Optimisation of the reasoning procedure for optimal human use, however, is still a problem that demands more research. Eventually, a broad open gap in XAI is the use of interpretabilities beyond using visualisation techniques. Future studies will demonstrate how to incorporate XAI into a broader optimisation mechanism in order to, e.g., boost the efficiency of the model and reduce the model complexity.

\section{Conclusion}

The recent confluence of large-scale annotated clinical databases, the innovation of deep learning approaches, open-source software packages, and inexpensive and rapidly increasing computing capacity and cloud storage has fuelled the recent exponential growth in AI. This foretells to change the landscape of medical practice in the near future. AI systems have specialised success in certain clinical activities that are more able to assess patient prognosis compared to doctors, and can help in surgical procedures. If deep learning models continue to advance, there is a growing chance that AI could revolutionise medical practice and redefine the role of clinicians in the process. Our mini-review has demonstrated the research trends towards the trustable AI or trustworthy AI, which promotes the XAI globally, and XAI methods in medicine and digital healthcare are highly in demand. Additionally, our two showcases have shown promising XAI results for the two most widely investigated classification and segmentation problems in medical image analysis. We can envisage further development of XAI in medicine and digital healthcare by integrating information fusion from cross-modalities imaging and non-imaging clinical data can be a stepping stone toward a more general acceptance of AI in clinical practice. Ultimately, the trustable AI will promote confidence and openness of its deployment in the clinical arena and also make it easier to comply with the legislation of the GDPR and regulations of the NHS$^X$ in the UK, CE-mark in the EU, FDA in the USA, and NMPA in China. 

\section*{Acknowledgement}

This work was supported in part by the European Research Council Innovative Medicines Initiative on Development of Therapeutics and Diagnostics Combatting Coronavirus Infections Award ‘DRAGON: rapiD and secuRe AI imaging based diaGnosis, stratification, fOllow-up, and preparedness for coronavirus paNdemics’ [H2020-JTI-IMI2 101005122], in part by the British Heart Foundation [PG/16/78/32402], in part by the AI for Health Imaging Award ‘CHAIMELEON: Accelerating the Lab to Market Transition of AI Tools for Cancer Management’ [H2020-SC1-FA-DTS-2019-1 952172], in part by the Hangzhou Economic and Technological Development Area Strategical Grant [Imperial Institute of Advanced Technology], in part by the Project of Shenzhen International Cooperation Foundation (GJHZ20180926165402083), and in part by the Clinical Research Project of Shenzhen Health and Family Planning Commission (SZLY2018018).

\bibliography{XAI_bib}

\end{document}